\documentclass[conference, a4paper]{IEEEtran}
\IEEEoverridecommandlockouts
\usepackage{cite}
\usepackage{amsmath,amssymb,amsfonts}
\usepackage{graphicx}
\usepackage{textcomp}
\usepackage{xcolor}
\usepackage{dsfont}
\usepackage{hyperref}
\usepackage{environ}
\usepackage{float} 
\usepackage{makecell}  % for \Xhline
\usepackage{booktabs}
\usepackage{mathtools}
\usepackage{cancel}
\usepackage{verbatim}
\usepackage{dsfont}
\usepackage{subfigure}
\usepackage{multirow}
\usepackage{bm}

\usepackage[ruled,linesnumbered]{algorithm2e}
\usepackage{float}

\newif\ifproofread

\newcommand{\changemarkertable}[1]{%
\ifproofread
{\color{blue}#1}%
\else
#1%
\fi
}

\makeatletter
\newcommand{\removelatexerror}{\let\@latex@error\@gobble}
\makeatother

\makeatletter
\newcommand\newtag[2]{#1\def\@currentlabel{#1}\label{#2}}
\makeatother

\usepackage{scalerel}
\usepackage{graphicx}

\def\BibTeX{{\rm B\kern-.05em{\sc i\kern-.025em b}\kern-.08em
    T\kern-.1667em\lower.7ex\hbox{E}\kern-.125emX}}
    
\begin{document}

%  True to highlight changes, false otherwise
\proofreadtrue

\title{NACHOS: Neural Architecture Search for Hardware Constrained Early Exit Neural Networks}

\author{\IEEEauthorblockN{Matteo Gambella, Jary Pomponi, Simone Scardapane, and Manuel Roveri}
\IEEEauthorblockA{Dipartimento di Elettronica, Informazione e Bioingegneria,\\
Politecnico di Milano, Milan, Italy\\
Sapienza Università di Roma, Rome, Italy \\
Email: \{matteo.gambella,manuel.roveri\}@polimi.it \\
Email: \{jary.pomponi,simone.scardapane\}@uniroma1.it}
}

\maketitle

%------
%il primo framework di NAS in letteratura che risolve i seguenti problemi in simultanea: (a) creare una architettura EE; (b) scegliere dove posizionare le EE tramite; (c) scegliere gli iperparametri di uscita; (d) ottimizzare in senso di Pareto efficienza ed accuratezza. Poi possiamo dire che il primo lo otteniamo estendendo il lavoro di prima, il (b)-(d) con nuovi termini di regolarizzazione. 
%------------------

%-----------------------------------------------------------------------------
% ABSTRACT
%-----------------------------------------------------------------------------
\begin{abstract}

Early Exit Neural Networks (EENNs) endow a standard Deep
Neural Network (DNN) with Early
Exit Classifiers (EECs), to provide predictions at intermediate
points of the processing when enough confidence in classification
is achieved. This leads to many benefits in terms of effectiveness
and efficiency. Currently, the design of EENNs is carried out manually by
experts, a complex and time-consuming task that requires accounting for many aspects, including the correct placement, the thresholding, and the computational overhead of the EECs. 
For this reason, the research is exploring
the use of Neural Architecture Search (NAS) to automatize the
design of EENNs. Currently, few comprehensive NAS solutions for EENNs have
been proposed in the literature, and a fully automated, joint design strategy taking into consideration both the backbone and the EECs remains an open problem. To this end, this
work presents Neural Architecture Search for Hardware
Constrained Early Exit Neural Networks (NACHOS), the first NAS framework for the design of optimal
EENNs satisfying constraints on the accuracy and the number
of Multiply and Accumulate (MAC) operations performed by the EENNs at
inference time. In particular, this provides the joint design of backbone and EECs to select a set of admissible (i.e., respecting the constraints)
Pareto Optimal Solutions in terms of best trade-off between the accuracy and number of MACs.
The results show that the models designed by NACHOS are competitive with the state-of-the-art EENNs.
Additionally, this work investigates the effectiveness of two novel regularization terms designed for the optimization of the auxiliary classifiers of the EENN.

\end{abstract}


\section{Introduction}%
\label{sec:introduction}

Early Exit Neural Networks (EENNs) are an emerging class of neural networks characterized by the presence of multiple early exits, called Early Exit Classifiers (EECs), placed at different points of the neural network architecture (backbone) to provide intermediate predictions when enough confidence in the classification is achieved~\cite{branchynet,scardapane_why_2020}. This leads to significant benefits in terms of effectiveness and efficiency, such as the mitigation of overfitting~\cite{overfitting}, overthinking~\cite{idk_cascades,shallowdeepnetworks}, and vanishing gradient phenomena~\cite{vanishinggradient}.

The motivating idea underlying EENNs is that the majority of input samples can be classified by resorting to smaller architectures~\cite{lottery_ticket_hypothesis}, even in very complex datasets such as ImageNet~\cite{imagenet}. Indeed, previous works have shown that the majority of input samples can be classified by the EENNs in their earlier stages, resulting in a significant reduction of the average inference time~\cite{branchynet}~\cite{scardapane_differentiable_2020}. This is particularly beneficial for embedded applications implemented by Internet-of-Things (IoT) units, embedded systems, and Edge computing units from the computation and energetic point of view.

For all these reasons, the research interest in designing effective and efficient EENNs is steadily growing with many solutions available in the literature enhancing state-of-the-art neural networks (e.g., Convolutional Neural Networks and Transformers) and addressing different tasks ranging from image classification and semantic segmentation to Natural Language Processing~\cite{surveyeenlp, badge, ee_har, ee_videoobjdetect, dynamicvit}. Currently, the design of EENNs is carried out manually by experts. Unfortunately, this is a very complex and time-consuming task since many aspects must be accounted for, such as how many EECs to consider, their placement, the setting of per-exit confidence thresholds~\cite{biglittleDNN}, and the calibration of the EECs~\cite{guo2017calibration}.

Neural Architecture Search (NAS) is a prominent research area~\cite{nasframework, geneticnas, fnnnas} aiming at automating the design of a neural network by exploring different architectural configurations given a neural network topology, a task to be accomplished, and a dataset used for training and validation so as to provide the optimal architecture. The majority of works in this field focus on the design of deep neural network architectures while taking into account a specific target hardware. This family of NAS solutions is called Hardware-Aware NAS~\cite{HardwareNAS}. An emerging direction of these NAS solutions regards the use of constraints in the NAS exploration~\cite{gambella_cnas_2022, micronas, hadas, li_automatic_2023}. Specifically, the works belonging to this research direction are able to consider both functional constraints (i.e., the type of operations that can be carried out in the neural network) and technological constraints (i.e., constraints on the computational and memory demand of the designed neural network in the exploration of the neural networks). The constraints are usually handled by a genetic algorithm applying a penalization to the fitness of the architectures not satisfying the hardware constraints. Interestingly, there are very few examples of NAS for EENNs in the current literature and these works are not able to provide an effective joint design of the backbone networks and the EECs~\cite{eexnas,esai,hadas,edanas}. These NAS solutions are commented in Section~\ref{sct:related_literature}.

\begin{comment}
Early exit aDAptive Neural Architecture Search (EDANAS)~\cite{edanas} is the first work aiming at developing a NAS framework for the joint design of backbones and EECs and able to handle hardware constraints. In particular, EDANAS designs the architecture of the backbone, the thresholds, and the placement of EECs. The algorithm has proven to compete with expert-designed early exit solutions.
\end{comment}

In this paper, we propose NACHOS, a shorthand for Neural Architecture Search for Hardware Constrained Early Exit Neural Networks, a novel NAS framework able to co-design the network architecture and the EECs and impose constraints on the classification accuracy and the number of Multiply and Accumulate (MAC) operations performed by the EENN at inference time. This solution, which is inspired by what is proposed in~\cite{edanas}, automatically designs and places the EECs in the candidate backbone of an EENN network according to a criterion regarding the number of MACs and tunes their thresholds accordingly.
Moreover, NACHOS introduces two regularization terms to enforce confidences around the operating point of the EECs to enhance the accuracy and enforce the respect of the constraint during the training of a candidate network.
Lastly, a novel figure of merit named $F_{\text{CM}}$ accounts for the number of MACs of an EENN and the related constraint adaptively during the NAS search.
Overall, NACHOS shows a significant improvement in terms of the effectiveness and efficiency of the designed EENNs with respect to baseline state-of-the-art approaches. We highlight that NACHOS falls into the family of Hardware-Aware NAS.

To sum up, the novel contributions introduced in this study
are the following:
\begin{itemize}
\item a novel NAS solution able to co-design the backbone and the EECs in EENNs satisfying user-defined constraints by automatically placing the EECs in the backbone of an EENN accounting for their computational complexity;
\item the introduction of a regularization term to ensure that the local certainties of the EECs follow their operating point in order to enhance the accuracy;
\item the use of a regularization approach to
increase the computational efficiency of the early-exit strategy and account for the constraint;
\item a novel figure of merit optimized by the genetic algorithm of the NAS to adaptively handle the constraint.
\end{itemize}

\begin{comment}
\item Switching off automatically some exits (to do)
\item SURROGATE of global gates (to do)
\end{comment}

The rest of this paper is organized as follows. Section~\ref{sct:related_literature} introduces the related works regarding EENNs designed by NAS solutions. Section~\ref{sct:background} provides the background. 
\begin{comment}
Section~\ref{subsct:problem_formulation} presents the problem formulation of NACHOS.
\end{comment}
Section~\ref{sct:NACHOS} introduces the NACHOS framework developed in this study, while in Section~\ref{sct:experiments} the experimental results aiming at evaluating the effectiveness of NACHOS are shown. Conclusions are finally drawn in Section~\ref{sct:conclusions}. To facilitate comparisons and reproducibility, the source code of NACHOS is released to the scientific community as a public repository.\footnote{https://github.com/AI-Tech-Research-Lab/CNAS}

\section{Related literature}
\label{sct:related_literature}

In the literature, many NAS solutions address explicitly the problem of energy efficiency of deep neural networks by optimizing both the accuracy in a given task and the hardware requirements (e.g., the memory and the computational demand). These works belong to a family of solutions called Hardware-Aware NAS~\cite{HardwareNAS}. In this field, a promising research direction regards the design of neural networks in constrained environments. These NAS solutions permit to impose constraints on the neural network exploration to focus the search in the region of feasible solutions accommodating for the hardware requirements ~\cite{gambella_cnas_2022, micronas, hadas, li_automatic_2023}. Among the Hardware-Aware NAS solutions, some recent works propose NAS solutions for optimizing the design of neural networks for EENNs, which are briefly reviewed here.

\begin{comment}
Traditionally, EENNs perform adaptive inference in a layer-wise fashion, i.e., placing the EECs between layers of the same network. Differently, other works are focused on a different mechanism of adaptive inference. The work~\cite{designadaptNN} is a NAS framework for neural networks performing early exit in a network-wise fashion meaning that EECs are placed between different networks in a cascade. S2DNAS~\cite{s2dnas} presents a NAS for neural networks performing adaptive inference in a channel-wise fashion, i.e. splitting the network along the channel width. 
\end{comment}

EExNAS~\cite{eexnas} is a first attempt to optimize an EENN with NAS. It accounts for only one EEC at a fixed position. Differently, ESAI~\cite{esai} is a NAS for EENNs in a Cloud setting. The EENN is organized into two sub-parts, i.e., a lighter one is processed locally and another heavier one on cloud. The early exit mechanism is used to decide where to send the samples to units (e.g., local or remote server) of a technological ecosystem. Both works operate on medical applications, the first focusing on Myocardial Infarction and Human Activity Recognition, and the second on the classification of dermoscopic images.
HADAS~\cite{hadas} presents a framework to design the backbone network, the EECs, and the dynamic voltage and frequency scaling settings, 
by a two-level optimization with two different genetic algorithms.
On the first level, HADAS selects the optimal backbone neural networks by sampling from a Once-For-All (OFA) supernet~\cite{ofa} and taking into consideration the hardware constraints, while on the second level, it enhances the chosen backbones with the EECs by automating the choice of their number and placement, trains the obtained EENNs, and then selects the optimal EENNs. The EENNs are trained by using the Knowledge Distillation (KD) approach. This NAS takes 2-3 GPU days on a cluster of 32
GPUs on CIFAR-100. 

A different approach suggests that the OFA can be enhanced and trained with early exits~\cite{sarti2023enhancing}. The work presents OFAv2, a novel OFA extending the OFAMobileNetv3 architecture by adding parallel blocks, additional dense skip connections, and implementing early exits. To support the redesign of the OFA, the training algorithm of the OFA, named Progressive Shrinking, has been expanded by adding two additional steps to handle the parallel blocks and the early exits, respectively. The NAS approach used in the experimental campaign relying on the novel OFA is evaluated on the Tiny Imagenet Dataset, a smaller version of Imagenet, which provides a significant increase in accuracy with respect to the original OFA architecture and maintains the computational efficiency. All in all, OFAV2 performs the search on a previously trained supernet specifically designed with multi-exits.

\begin{comment}
Currently, the NAS solutions for EENNs are very limited in jointly designing the backbones and the EECs and don't take into account their computational demand. 
\end{comment}
Currently, there are only few NAS solutions that jointly design the backbones and the EECs while taking into account the computational demand.
EDANAS~\cite{edanas} is a framework for the automatic design of both the backbone architecture and the parameters managing the early exit mechanism of the EENNs so as to optimize both the accuracy of the classification tasks and the computational demand.
The problem addressed by EDANAS is formulated as a joint optimization problem consisting in selecting the architecture of an EENN aiming at providing a high accuracy on a given task and reducing the MACs associated with the EE by using the genetic algorithm NSGA-II~\cite{nsga-II}. Similarly, HarvNet~\cite{harvnet} addresses the joint design of backbone and EECs while meeting memory and energy constraints in dynamic harvesting environments. It relies on reinforcement learning to adjust the confidence threshold of each exit to maximize accuracy by considering the device's runtime factors. This NAS takes 47 GPU days on CIFAR-10, since they integrate the runtime aspects of dynamic input mappings and dynamic harvesting environments in the design of NNs respectively.

\begin{comment}
\changemarkertable{We emphasize the discrimination of our work with previous ones. EExNAS and ESAI operate on medical applications, the first focusing on Myocardial Infarction and Human Activity Recognition, and the second on the classification of dermoscopic images. HADAS and HarvNet work on computer vision applications but require high hardware requirements to work, the first takes 2-3 GPU days on a cluster of 32 GPUs on CIFAR-100 while the second takes 47 GPU days on CIFAR-10, since they integrate the runtime aspects of dynamic input mappings and dynamic harvesting environments in the design of NNs respectively. OFAV2 performs the search on a previously trained supernet specifically designed with multi-exits. Our work shows a hardware-aware NAS for EENNs on computer vision applications working on top of a standard single-exit OFA supernet independent of input mappings and the environment with a search time of 2 GPU days on a single GPU (NVIDIA A40). }
\end{comment}

Lastly, we emphasize that our work shows a hardware-aware NAS for EENNs on computer vision applications working on top of a standard single-exit OFA supernet independent of input mappings and the environment with a search time of 2 GPU days on a single GPU (NVIDIA A40). This overcomes some of the limitations of the previous works such as limited application scenarios, specific design of the supernet, and high computational requirements.

\section{Background}
\label{sct:background}

\subsection{Early Exit Neural Networks}
\label{subsct:eenn}
We can see a neural network as a composition of $b$  sequential blocks: 

\begin{equation}
    f(x) = \mathrm{B}_{b} \circ \mathrm{B}_{b-1} \circ \cdots \circ \mathrm{B}_{1}(x) 
\end{equation}

\noindent where $\circ$ denotes function composition. This is a generic definition which is agnostic about the internal topology of a block, and that applies to most neural networks. Moreover, we can apply a classification layer to each block i-th of the model, which leads to multiple prediction functions:

\begin{equation}
\label{eq:composition}
    f_i(x) = \mathrm{C}_{i} \circ \left( \mathrm{B}_{i} \circ \mathrm{B}_{i-1} \circ \cdots \circ  \mathrm{B}_{1}(x) \right)
\end{equation}

This formulation leads to an early exit model which is capable of classifying input samples at any stage of the computation. The result is a sequence of predictions for a given input sample, and different exits can mistake different subsets of the dataset. Thus, combining them can improve the overall performance, or reduce the computational time required to calculate the prediction associated with a sample, under a given measure of certainty, which is used to halt the computation in a given branch, if the prediction is good enough. 

A common way to train such models is by jointly updating all the exits: 
\begin{equation}
    \label{eq:joint}
    L_{\text{joint}}(x, y) = l(f_b(x), y) + \sum_{i=1}^{b-1} \lambda_i \cdot \underbrace{\left[ l(f_i(x), y) \right]}_{\text{$i$-th branch loss}} \,,
\end{equation}
\noindent where $l(\cdot, \cdot)$ is the loss (e.g. cross-entropy loss), and $\lambda_i$ the weight associated to the \textit{i}-th exit's loss. Other training approaches are \textit{Layer-wise Training} (LT), aiming at separately optimizing each layer in the NN architecture~\cite{marquez_deep_2018}, and \textit{Knowledge Distillation (KD)} where the backbone network is trained separately from the EECs and then the EECs are trained on top of it by relying on KD~\cite{knowledgedistillation}.

After training the model using the aforementioned approach, we can take advantage of the formulation proposed in Eq. \ref{eq:composition} to select the earliest feasible exit which is assumed to output the correct prediction. This leads to many benefits such as the reduction of the inference time~\cite{branchynet}~\cite{scardapane_differentiable_2020}, overfitting~\cite{overfitting} and overthinking~\cite{branchynet}. To do this we can use different early exit mechanisms, such as entropy, proposed in \cite{branchynet}, the largest value of the softmax, or the so-called Score Margin (SM)~\cite{biglittleDNN}~\cite{designadaptNN}, which is the difference between the largest and the second largest value on the softmax on the prediction. These mechanisms are simple to implement, but they introduce a potential mismatch between training and test behaviours, since the early exit mechanism is not optimized on the training data.

In this paper we follow the alternative formulation proposed in \cite{pomponieenn}, which, using a hierarchical differentiable mechanism, includes the early exit mechanism into the exit itself. To do that, along with the EEC $\mathrm{C}_i$, we define also a confidence score $c_i(\cdot)$ which outputs a scalar in $[0, 1]$, which follows the same composition rule proposed in Equation \eqref{eq:composition}. So we have that, at a given branch $i$, we have:

\begin{equation}
    \label{eq:confidence}
    f^c_i(x) =
        \begin{bmatrix}
    f_i(x) \\
    c_i(x) 
    \end{bmatrix} \,.
\end{equation}

The idea is that the confidence score should output a value which measures how much the exit is confident of its prediction. We recursively combine these confidences to produce a cumulative confidence score defined as:

\begin{comment}
\begin{equation}
\label{eq:recursive_formulation}
    \overline{y}_e(x) = h_i(x) \cdot f_i(x) + (1-h_i(x)) \cdot \overline{y}_{i+1}(x),
\end{equation}

with $h_b(\cdot) = 1$ to force the computation to halt when reaching the final exit. To take advantage of this formulation during the inference phase, we recursively combine the halting heads as

\end{comment}

\begin{equation}
    c_i^r(x) = c_i(x)  \prod_{k=1}^{i-1} (1 - c_k(x)) \,,
    \label{eq:branch_process}
\end{equation}

\noindent with $c_b(\cdot) = 1$ to force the computation to halt when reaching the final exit. To produce the final output, we integrate the just defined confidence scores into the forward step:

\begin{equation}
    % \label{eq:confidence}
    \overline{f}_k(x) = \sum_{i=1}^{k} c_i^r(x) f_i(x)\,.
\end{equation}

The overall model is trained to reduce the following loss

\begin{equation}
\label{eq:recursive_loss}
    L_{\text{acc}}(x, y) = l(f_b(x), y) + l(\overline{f}_{b-1}(x), y) 
\end{equation}

We divide the loss into two parts to avoid the intermediate branches, the second term, overshadowing the final output, the first term, which can be small if the model is highly confident in the early stage of the computation.  This training approach allows us to halt the inference process at a given exit $i$ when the following criterion is met:

\begin{equation}
 \label{eq:confidencethreshold}
  c_i(x) \prod_{k=1}^{i-1} (1 - c_k(x))  \ge \epsilon,
\end{equation}

\noindent where $\epsilon \in [0, 1]$ is a confidence threshold. In order to give representative values, the early exits must be regularized. To do that, the following regularization loss, which is added to Equation \eqref{eq:recursive_loss}, is employed for each exit $i < b$:

\begin{align}
    \label{eq:reg}
    \mathcal{R}_{i}(x, y) = &-c_i(x) \log(\mathds{1}(y, \overline{f}_i(x)) \nonumber \\ 
    &- (1 - c_i(x)) \log(\mathds{1}(y, \overline{f}_i(x)) \,,
\end{align}

\noindent with $\mathds{1}(y, \overline{f}_i(x))$ a function that returns 1 if the $i$-th exit's prediction is correct with respect to the true label y, otherwise 0. This function is not differentiable, and to overcome this we calculate these ground truth values before the training step associated to the sample $x$,  without interfering with the gradient computations. Moreover, a sampling trick is used to ensure that the resulting model is not overconfident, and we remand to the original paper for further details. 

\subsection{Neural Architecture Search}
\label{subsct:nas}
The NAS solutions can be categorized according to three different dimensions~\cite{nas_survey_components}. 

The first dimension is the \textit{Search Space}, which defines the architecture to be represented. The most straightforward \textit{Search Space} is \textit{entire-structured} which is represented layer-wise with one node representing a layer. Motivated by hand-crafted architectures consisting of repeated motifs~\cite{mobilenetv3}, the \textit{cell-based Search Space} defines each node as a cell (also dubbed block), a group of layers. Other types of search space show cells with different levels (hierarchical) or identity transformations between the layers (morphism-based).

The second dimension is the \textit{Search Strategy}, which establishes how to explore the search space. Common strategies are based on reinforcement learning (RL)~\cite{rlintro,rlsurvey}, gradient optimization~\cite{liu_darts_2019}, and evolutionary algorithms (EAs)~\cite{eabook,eaoptimization}. The latter, in particular the genetic methods, represents the most popular choice~\cite{surveyevonas}. 
\begin{comment}
A typical genetic algorithm deployed in many NAS solutions is NSGA-II~\cite{nsga-II}, which generates offspring by using a specific type of crossover and mutation and then selects the next generation showing the best fitness according to nondominated-sorting and crowding distance comparison.
\end{comment}
The Non-dominated Sorting Genetic Algorithm II (NSGA-II) is widely used in NAS solutions thanks to its ability to efficiently explore the search space~\cite{nsga-II}. It generates offspring through crossover and mutation operations, and then applies a non-dominated sorting procedure combined with a crowding distance mechanism to select the most optimal individuals for the next generation. This approach ensures that the selected architectures are not only characterized by high fitness but also maintain a diverse set of solutions, helping to avoid premature convergence. Moreover, it is a multi-objective algorithm, i.e., being able to optimize jointly different figures of merit, and it is able to effectively handle constraints~\cite{nsgaiiconstraints}. The constraints are handled by redefining the objective function~\cite{eaconstraints}, and the simplest scheme works as follows:
\begin{equation}
\label{eq:constraints}
\phi(x) = f(x)+pG(x)
\end{equation}
where $\phi(x)$ is the joint objective, \textit{f(x)} is the original objective without constraints, \textit{p} is the penalty term and \textit{G(x)} is a function deciding whether to apply the penalty by accounting for the constraints. The main idea is that the application of a penalty results in a bad fitness in the genetic algorithm and precisely in a bad ranking in the case of NSGA-II proportional to the constraint violation. The penalty may be static, i.e., a constant value, or dynamic, i.e. adapted during the evolutionary process~\cite{eaconstraints}.

The third dimension is the \textit{Performance Estimation Strategy}, which refers to the process of estimating the performance to be optimized. The purpose of this family of methods is to avoid training every single neural network architecture in the search space which would be prohibitive in terms of computational time. Weight sharing is a popular estimation strategy, and in particular, Once-For-All (OFA)~\cite{ofa} represents a state-of-the-art. OFA trains a supernet comprehensive of many network configurations once and evaluates the candidate networks by fine-tuning from the supernet weights during the search of the NAS. The training of the OFA supernet is performed by a technique called \textit{progressive shrinking}, which first trains the largest neural network with maximum depth, width, and kernel size, then progressively fine-tunes the OFA network to support smaller sub-networks that share weights with the larger ones. While OFA provides three variants of CNN-based supernets, inspired by MobileNetV3, ProxylessNAS, and Resnet50  respectively, NASViT proposed a Vision Transformer-based supernet \cite{gong2022nasvit}. The choice of the supernet accounts for the different desired accuracy-efficiency trade-off in a target scenario. Another solution is the use of surrogate models such as a Gaussian Process~\cite{gp} and Radial Basis Function~\cite{rbf}. For instance, MSuNAS~\cite{msunas} proposes a method called \textit{adaptive-switching} which selects the best accuracy predictor among four possible surrogate models according to a correlation metric named Kendall’s Tau~\cite{kendalltau} at each iteration of the NAS process.  Moreover, another technique called \textit{weight inheritance} (network morphism)~\cite{networkmorphism} evaluates the performance of a network obtained by an identity transformation (morphism) on a parent model by inheriting the weights of the parent model, i.e., being initialized with the same weights.

\begin{comment}
\textbf{OFA}

\textbf{rewrite this part}

Actually, many Hardware-Aware NAS works rely on Once-For-All (OFA) supernets~\cite{bossnas,gambella_cnas_2022,edanas,liu_darts_2019,msunas}. 
The key idea is to train a single large network with a generic architecture and then fine-tune, truncate, and adapt it for different hardware platforms and resource constraints, while keeping the initial macro-architecture fixed. This goal is achieved through a training technique the authors called
Progressive Shrinking (PS), which consists of many phases to progressively fine-tune smaller and smaller networks from within the supernet.
Then, during the search, the networks to be evaluated are trained by fine-tuning from the supernet weights, hence increasing the computational efficiency of the NAS. 

Notably, the training of the supernet overcomes the \textit{multi-model forgetting} problem~\cite{multi_model_forgetting} that occurs when sequentially training multiple deep networks with partially-shared parameters and performance of previously trained networks degrade due to the overwrite of shared parameters. Compared to other hardware-aware NAS techniques, this is considered to be one of the less computationally expensive.
\end{comment}

\section{The proposed NACHOS}
\label{sct:NACHOS}

This section introduces and details the presented NACHOS. More specifically, 
Section~\ref{subsct:problem_formulation} provides the problem formulation of NACHOS,
Section~\ref{subsct:overallview} provides an overall description of NACHOS,
Section~\ref{subsct:eecplacement} details the automatic placement of the EECs along the backbone,
\begin{comment}
Section~\ref{subsct:globalgate} explains the automatic branch selection performed by the novel global gate.
\end{comment}
Section~\ref{subsct:training} describes briefly the training phase by detailing the introduced regularization terms and the inference phase of the EENNs. Finally, Section~\ref{subsct:cmacs} introduces a novel figure of merit handling for the number of MACs of the EENN and the corresponding constraint enforced. To improve the readability, Table \ref{tab:terminology} contains all symbols we use to define the problem and our proposal.

\subsection{Problem formulation}
\label{subsct:problem_formulation}

The problem of NACHOS consists in selecting an EENN optimizing the classification accuracy and number of MAC operations performed by the EENNs at inference time while respecting constraints on these two figures of merit. Formally, NACHOS can be defined as the following constrained optimization problem:
\begin{align}
\label{eq:constrained_problem}
    \text{minimize }  &\mathcal{G} \left ( 
    F_A(\widetilde{x}),F_M(\widetilde{x}) \right ) \nonumber \\
    \text{s. t. } & F_A(\widetilde{x}) > \overline{F}_A, \\
    & F_M(\widetilde{x}) < \overline{F}_M, \nonumber \\
                      &\widetilde{x} \in \Omega_{\widetilde{x}} \nonumber
\end{align}
where $\mathcal{G}$ is a multi-objective optimization function, $\widetilde{x}$ and $\Omega_{\widetilde{x}}$ represent a candidate neural network architecture and the search space of the neural network exploration, respectively, $F_A(\widetilde{x})$ and $\overline{F}_A$ are the metric computing the top1 accuracy and the corresponding constraint, $F_M(\widetilde{x})$ and $\overline{F}_M$ are, respectively, the metric computing the number of MACs of an EENN and the corresponding constraint. The optimization problem described in Equation \eqref{eq:constrained_problem} will be tackled by the proposed NACHOS framework that is introduced in the next Section.

\subsection{The overall view of NACHOS}
\label{subsct:overallview}
\begin{figure*}[t]%
	\centering
	\includegraphics[width=0.9\linewidth]{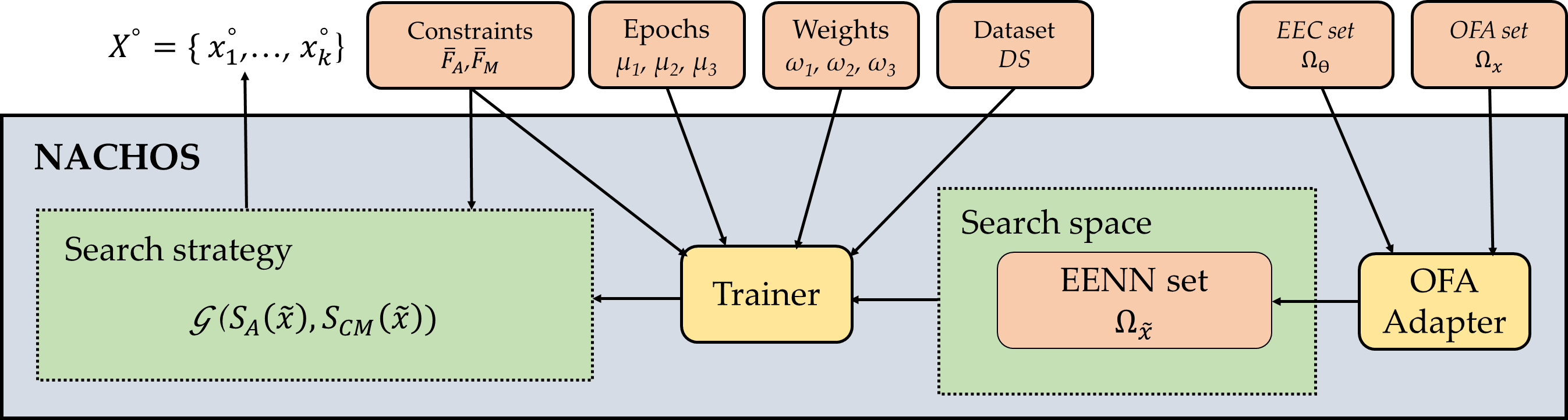}
	\caption{The proposed NACHOS framework, which is composed of an OFA Adapter, Trainer and a Search Strategy module.} 
	\label{fig:NACHOS_scheme}
\end{figure*}

\begin{figure}[t]%
	\centering

	\includegraphics[width=0.55\linewidth]{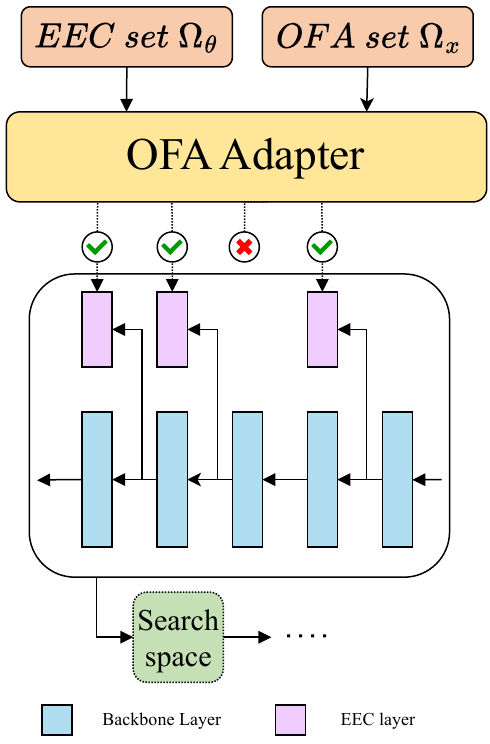}
	\caption{The process of creating an EENN and how such models are used to populate the searching space. The best model is searched within this space. For further details, please refer to Section \ref{subsct:eecplacement}.}
 
	\label{fig:ofa_eenn}
\end{figure}
% \begin{figure*}[htp]
%   \centering
%   \subfigure[The proposed NACHOS framework, which is composed of an OFA Adapter, Trainer and a Search Strategy module.]{	\includegraphics[width=0.58\linewidth]{Images/CBNAS_scheme.png}
% }\quad
%   \subfigure[random caption 2]{\includegraphics[scale=0.4]{Images/NACHOS_eenn_scheme.pdf}}
% \end{figure*}
This section introduces the proposed NACHOS, which is a NAS solution able to jointly optimize the design of EENN and its EECs.

An overall description of the NACHOS framework is given in Fig.~\ref{fig:NACHOS_scheme}. In more detail, NACHOS receives as input a dataset $\mathcal{DS}$, comprising a training set (used to train the candidate networks) and a validation set (used to validate the candidate networks), the set of constraints $\{\overline{F}_A,\overline{F}_M\}$ enforced in the NAS exploration, the set $\{\mu_1,\mu_2,\mu_3\}$ consisting of the number of epochs for the steps of the training process, the set $\{\omega_1,\omega_2,\omega_3\}$ consisting of the weights of the training loss function \textit{L}, the set $\Omega_x$ containing the candidate backbones \textit{x} of the EENNs sampled by the OFA supernet, and the set $\Omega_{\theta}$ containing the hyperparameters $\theta$ of the EECs.

In more detail, the \textit{OFA Adapter} module receives as inputs $\Omega_{\theta}$ and $\Omega_x$ and enhances each $x$, candidate network of the OFA, by placing EECs along the backbone in the positions given by the configuration $\theta$ sampled from the set $\Omega_{\theta}$. The output is a new set $\Omega_{\widetilde{x}}$ consisting of EENNs represented with $\widetilde{x}$. Moreover, the \textit{OFA Adapter} automatically designs the architecture of the EECs as introduced in Section \ref{subsct:eecplacement}.

The \textit{Trainer} module takes in input the EENNs from the set $\Omega_{\widetilde{x}}$ and trains the EENNs on the $\mathcal{DS}$ dataset. Each step of the training process is run for $\mu_i$ epochs with $i$ being the number of the step. Each component of the loss function is weighted by a weight $\omega_i$. After the training, the EENNs are evaluated by inference with a set of thresholds, and the best configuration of the thresholds is chosen according to a criterion accounting for the set of constraints. The output is a set, named the \textit{archive}, consisting of the tuples $<\widetilde{x},F_A(\widetilde{x}),F_M(\widetilde{x})>$, representing the configuration of the EENN and the related figure of merits $F_A(\widetilde{x})$ and $F_M(\widetilde{x})$ on accuracy and computation, respectively.
Before starting the search, a representative set of the OFA $\Omega_{\widetilde{x}}$, is sampled. The number $N_{start}$ of EENNs sampled is a parameter chosen by the user (usually $N_{start}=100$). This means that the cardinality of the archive is initially equal to $N_{start}$.

In NACHOS, the optimization problem defined 
in Eq. \eqref{eq:constrained_problem} is reformulated as follows by handling the constraints implicitly:
\begin{align}
\label{eq:NACHOS_problem}
    \text{minimize }  &\mathcal{G} \left ( \mathcal{S_{A}}(\widetilde{x}), 
    \mathcal{S_{CM}}(\widetilde{x}) \right ) \\
    \text{s. t. }  & \widetilde{x} \in \Omega_{\widetilde{x}}
    \nonumber
\end{align}
where $\mathcal{S_{A}}(\widetilde{x})$ is the predicted classification accuracy of $\widetilde{x}$ computed by the accuracy predictor introduced in MSuNAS~\cite{msunas} and $\mathcal{S_{CM}}(\widetilde{x})$ is $F_{CM}(\widetilde{x})$, introduced in Eq.~\eqref{eq:cmacs}, computed on $\mathcal{S_{M}}(\widetilde{x})$, the predicted $F_M(\widetilde{x})$ taken by the surrogate model (i.e., the MACs predictor).

The fundamental part of NACHOS is the \textit{Search Strategy} module. It adopts the genetic algorithm NSGA-II~\cite{nsga-II} to solve the bi-objective problem introduced in Equation \eqref{eq:NACHOS_problem}, by jointly optimizing the objectives $\mathcal{S}_{A}(\widetilde{x})$ and $\mathcal{S}_{CM}(\widetilde{x})$  on the dataset $\mathcal{DS}$, and by evaluating network architectures candidates from $\Omega_{\widetilde{x}}$. 
The search process is iterative: at each iteration, the two surrogates model, computing $\mathcal{S}_{A}(\widetilde{x})$ and $\mathcal{S}_{M}(\widetilde{x})$, respectively, are chosen employing a mechanism called \textit{adaptive-switching}, the same method used by MSuNAS~\cite{msunas}, which selects the best surrogate model according to a correlation metric (i.e., Kendall's Tau~\cite{kendalltau}). The surrogate models are trained by using the \textit{archive} as the dataset. Then, a ranking of the candidate networks, based on $\mathcal{S}_{A}(\widetilde{x})$ and $\mathcal{S}_{CM}(\widetilde{x})$, is computed and a new set of candidates is obtained by NSGA-II and forwarded to the \textit{Trainer}. The \textit{Trainer} updates the \textit{archive}, which becomes available for evaluation in the next iteration. The search ends if a number of iterations $N_{iter}$ is reached or the hypervolume of the Pareto front is not changing for $N_{tol}$ iterations (convergence).

At the end of the search, NACHOS returns the set of the $k$ EENN architectures $X^\circ = \{ x_1^\circ, \ldots, x_k^\circ \}$ characterized by the best trade-off among the objectives, being $k$ a value specified by the user. We emphasize that the output of NACHOS is a set of networks enhanced with EECs, while the OFA given in input consists in regular neural networks (i.e., no EECs).

\subsection{The OFA Adapter module: automatic placement of EECs}
\label{subsct:eecplacement}

\begin{table}[t!]
\centering
\caption{List of the main symbols used in the experiments}
\label{tab:terminology}
\scalebox{0.8}{
\begin{tabular}{@{}c|c@{}}
\toprule
\textit{Symbol}  &  \textit{Meaning} \\ \midrule
$\Omega_x$ & Set of OFA backbones.\\ 
\midrule
$\Omega_{\theta}$ & Set of EEC hyperparameters.\\ 
\midrule
$\Omega_{\widetilde{x}}$  & Set of EENNs built on top of $\Omega_x$ and $\Omega_{\theta}$.\\ 
\midrule
$x$  & Neural network architecture (backbone) sampled by $\Omega_x$.\\ 
\midrule
$\theta$  & The configuration of hyperparameters sampled by the EEC set.\\ 
\midrule
$\epsilon$ & The set of confidence thresholds. \\
\midrule
$\widetilde{x}$  & EENN sampled by $\Omega_{\widetilde{x}}$.\\ 
\midrule
$F_A(\widetilde{x})$  & The top1 accuracy of $\widetilde{x}$.\\ 
\midrule
$F_M(\widetilde{x})$  & The number of MACs of $\widetilde{x}$.\\ 
\midrule
$F_{CM}(\widetilde{x})$  & Novel formulation of $F_M(\widetilde{x})$ endowed by handling the constraint $\overline{F}_M$.\\ 
\midrule
$\overline{F}_A$ & The constraint on the figure of merit $F_A(\widetilde{x})$. \\
\midrule
$\overline{F}_M$ & The constraint on the figure of merit $F_M(\widetilde{x})$. \\
\midrule
$S_A(\widetilde{x})$  & The top1 accuracy of $\widetilde{x}$ predicted by the surrogate model.\\
\midrule
$S_M(\widetilde{x})$  & The number of MACs of $\widetilde{x}$ predicted by the surrogate model.\\
\midrule
$S_{CM}(\widetilde{x})$  & The value of $F_{CM}(\widetilde{x})$ computed on $S_M(\widetilde{x})$. \\
\midrule
$U_i(\widetilde{x})$  & The ratio of samples exiting at the $i^{th}$ of $\widetilde{x}$. \\
\midrule
$\gamma_i(\widetilde{x})$  & The computational cost of exiting at the $i^{th}$ of $\widetilde{x}$. \\
\midrule
$\mu_{i}$  & The number of epochs of the $i^{th}$ phase of the training process of an EENN.\\
\midrule
$\omega_{i}$ & The weight of the $i^{th}$ component of the loss training function. \\
\midrule
\end{tabular}
}
\end{table}

This section details the OFA Adapter module whose main goal is the automatic placement of the EECs along the backbone by taking into account the computational cost of the EECs.

We define $\gamma_i$ as the computational cost of exiting at $i^{th}$ exit. This cost can be computed as the sum of the computational cost of the backbone up to the $i^{th}$ exit and the computational cost of the $i^{th}$ EEC. In our framework, the computational cost $\gamma_i(\widetilde{x})$ is computed as: 
% NOTE: manca una figura di EENNs
\begin{equation}
\gamma_i(\widetilde{x}) = F_M(x_i) + F_M(EEC_i)
\end{equation}
where $x_i$ is the backbone $x$ up to the $i^{th}$ exit and $EEC_i$ is the $i^{th}$ EEC.

The basic design of these EECs is a sequence of ReLU activation functions and Convolutional layers followed by a flattening operation and a Fully-Connected layer~\cite{pomponieenn}.

The placement of the EECs is given by sampling from $\Omega_{\theta}$ a binary encoding configuration $\theta$ of length $\overline{B}-1$ being $\overline{B}$ a user-defined parameter representing the maximum number of EECs, distributed along the backbone equidistantly and including the final classifier. The uniform placement ensures that the computational complexity is distributed uniformly along the backbone since the layers have similar computational complexity. For instance, with $\overline{B}=5$, a sample $\theta=[1,0,1,0]$ extracted from $\Omega_{\theta}$ means that EECs are placed at $1/5$ and $3/5$ of the backbone. 
\begin{comment}
Note that the final classifier of the backbone network is not considered, since an EEC is added to the last exit of the EENN by this procedure.
\end{comment}

The steps of the placement of the EECs along the backbone are shown in Algorithm \ref{alg:ofaadapter}, while the process is visually depicted in Figure \ref{fig:ofa_eenn}. In this Algorithm, the EECs are placed along the backbone $x$ according to the configuration $\theta$ (Line 2). The vector of computational costs $\gamma$ is computed on $x$ and EECs (Line 3). In the \textit{For} statement each EEC is scanned and the condition $\gamma_i \le \gamma_{i+1}$ is checked (Lines 5-9). The last exit is the only one fixed a priori (i.e., basic design by default). If an EEC violates this condition, a \textit{Max Pooling} layer is added through the operator AddMaxPooling$(\cdot)$, where the window of the \textit{Max Pooling} layer is the minimum s.t. this EEC satisfies the condition. Finally, the designed EECs are attached to the backbone leading to the creation of an EENN (Line 10) and the EENN is added to $\Omega_{\widetilde{x}}$ (Line 11).

Other works propose an alternative condition to handle the complexity of the EECs~\cite{scardapane_why_2020}~\cite{baccarelli_optimized_2020}. This is applied after training when thresholds and percentage of samples exiting every EEC are known. Differently, our criterion is applied before training and is agnostic of the thresholds and exit ratios.

An OFA Adapter is present also in the EDANAS architecture. However, in the latter work, the structure of the EECs is fixed a priori and the module does not account for their complexity.

\begin{comment}
\subsection{Global Gate (1): Scalar values}
\label{subsct:globalgate}
This section considers the problem of selecting automatically a subset of all possible early exits to keep after training in an end-to-end way. 

This method is very straightforward. We add before each classifier $C_i$ a small global gate $\sigma(b_i)$, where $b_i$ is a trainable scalar value independent of any input. The gate is then activated if the global gate is close to one and if the confidence for the current input is also high:

\begin{equation}
\widetilde{y}_i={\sigma(b_i)}\left[c_iy_i +(1-c_i)\tilde{y}_{i+1}  \right] +{(1-\sigma(b_i))}\widetilde{y}_{i+1} 
\end{equation}

which simplifies to (double check):

\begin{equation}
\widetilde{y}_i=\sigma(b_i)c_iy_i + (1-\sigma(b_i)c_i)\widetilde{y}_{i+1}
\end{equation}

i.e., the gate acts as a global multiplicative factor on the local confidences for the current input. It is possible to select a subset of exits by enforcing an $\ell_1$ regularization on the vector $\mathbf{b}$ collecting all scalars $b_i$.
\begin{equation}
L_{gate}=\lVert \sigma(b_i) \rVert_1
\end{equation}

\end{comment}

% \algcaption{Ciao}

% \begin{figure} % Use [H] to force exact placement
%   % \begin{minipage}{\linewidth}
%   \removelatexerror
    \begin{algorithm}
    % \algcaption{Ciao}
    \caption{
  OFA Adapter update: the update of the $\Omega_{\widetilde{x}}$ performed by the OFA adapter}
    %\LinesNumbered 
  \KwIn{ backbone $x$, configuration $\theta$, the set of the EENNs $\Omega_{\widetilde{x}}$
  }
  \label{alg:ofaadapter}
 % Remove the final classifier from $x$\; 
  Design a set of $EECs$ based on $x$ and $\theta$\; 
  Initialize a variable $B$ with the number of $EECs$\;
 Define the vector $\gamma$ of computational costs based on $x$ and $EECs$\;
 \For{$i = B-1$ to $0$}{
\If{$\gamma_i > \gamma_{i+1}$}{
    $EECs_i \gets \text{AddMaxPooling}(EECs_i,\gamma_{i+1})$\;
    }
}
Attach $EECs$ to the backbone $x$ to create a EENN $\widetilde{x}$\;
Add $\widetilde{x}$ to $\Omega_{\widetilde{x}}$\;
 
\textbf{return} $\Omega_{\widetilde{x}}$ \;
      % \caption{OFA Adapter update}
    % \end{algorithm}
  % \end{minipage}
  % \label{fig:algo1}
  \end{algorithm}
% \end{figure}

\subsection{Training and inference of the EENNs in the Trainer module}
\label{subsct:training}

This section explains how the training of the EENN and the inference are performed in the Trainer module. The training is performed in an end-to-end way optimizing the loss consisting of an accuracy loss and the regularization terms regarding the computational constraint and the operating point of the EECs.  

% \begin{figure} % Use [H] to force exact placement
%   % \begin{minipage}{\linewidth}
%   \removelatexerror
    \begin{algorithm}
    \label{alg:eenntraining}
    
      \caption{Training of an EENN}
    %\LinesNumbered 
\KwIn{ EENN $\widetilde{x}$, dataset $DS$, constraints $\{\overline{F}_A, \overline{F}_M\}$, epochs $\{\mu_1, \mu_2, \mu_3\}$, weights $\{\omega_1, \omega_2, \omega_3\}$
  }
  Split the dataset $DS$ into a training and a validation set\;
  Get the backbone $x$ from $\widetilde{x}$ with the pretrained weights from the OFA supernet\; 
  Perform fine-tuning on $x$ for $\mu_1$ epochs\; 
  Define the vector of computational costs $\gamma$ from $\widetilde{x}$\;
  Train $\widetilde{x}$ applying a regularization $L_{cost}$ based on $\gamma$ and the confidences $c_i$\ for $\mu_2$ epochs\;
  \For{$\mu_3$ epochs}{
      Extract a set from the training set (Support Set)\;
      Compute the average accuracies per class on the Support Set\;
      For each batch, applies the regularization $L_{cost}$ and the regularization $L_{peak}$ based on the average accuracies and the confidences $c_i$ computed on the batch samples\;
  }
  Set $\widetilde{x}$ with the weights for the best validation loss\;
  \textbf{return} $\widetilde{x}$ \;
      % \caption{Training of an EENN}
    % \end{algorithm}
  % \end{minipage}
  % \label{fig:algo2}
% \end{figure}
\end{algorithm}

The training performed by the Trainer module is shown in Algorithm \ref{alg:eenntraining}. Firstly, the Trainer extracts the training and validation set from the dataset $DS$ (Line 1). Then, it performs fine-tuning on the backbone of the EENN for a number of epochs $\mu_1$ starting from the pre-trained weights of the OFA supernet (Lines 2-3). Once trained the backbone, the Trainer trains the EENN with the soft exit mechanism presented in Section ~\ref{subsct:eenn} with two regularization terms, $L_{cost}$ and $L_{peak}$. In particular, we can distinguish between a phase where only a regularization $L_{cost}$ on computational costs is applied for a $\mu_2$ number of epochs (Lines 4-5) and a second phase where an additional regularization $L_{peak}$ for enhancing the accuracy is applied lasting $\mu_3$ number of epochs. This is done to trigger the latter regularization only after the EECs have been trained for a sufficient number of epochs. (Lines 6-10). We then select the weights corresponding to the best loss computed on the validation set (Line 11).

The overall train loss function is the following:
\begin{equation}
\label{eq:trainlossNACHOS}
L = \omega_1 * L_{acc} + \omega_2 * L_{cost} + \omega_3 * L_{peak} 
\end{equation}
where:
\begin{itemize}
    \item $L_{acc}$ is the accuracy term shown in Equation \eqref{eq:recursive_loss}
    \item $L_{cost}$ is the regularization term to enforce the constraint on the complexity
    shown in Equation \eqref{eq:costloss}
    \item $L_{peak}$ is the regularization term introduced to ensure that the confidences of the EECs follow their operating point shown in Equation \eqref{eq:suploss}
\end{itemize}

The weights $\omega_1$, $\omega_2$ and $\omega_3$, given by the user (see \textit{Weights} in Fig.~\ref{fig:NACHOS_scheme}), are fixed in our setting but they could be learned during training to provide a simple way of forcing a desired trade-off between accuracy and energy-efficiency. We emphasize that the number of epochs of each step of the training is provided by the user as shown in Fig.~\ref{fig:NACHOS_scheme}. $\omega_3$ is set to zero to avoid applying the $L_{peak}$ during the step following the fine-tuning. 
In the rest of the Section, we explain the aforementioned regularization terms.

First, we examine the regularization to enforce the constraint on the complexity term. We aim at training more the fraction of the EENN not violating the constraint. As introduced in EDANAS~\cite{edanas}, the metric $F_M(\widetilde{x})$ can be computed as the sum of $\gamma_i$ weighted by the utilization of the corresponding $EEC_i$ and can be computed as follows:
\begin{align}
\label{eq:ada_macs}
F_M(\widetilde{x})=\sum_{i=1}^{B+1} \gamma_i(\widetilde{x})U_i(\widetilde{x})
\end{align}
where $U_i$ accounts for the percentage of samples classified by the i-th classifier. For simplicity, we use the operator $F_M(\cdot)$ to compute the MACs of both EENNs and neural networks without EECs, where $U_{B+1}(x)=1$ and $B=0$.

Following the considerations of a previous work~\cite{pomponieenn},
we can define $\widetilde{\gamma_i}(\widetilde{x})$ as the recursive definition of the previously defined computational cost $\gamma_i(\widetilde{x})$ in an analogous way to the soft exit mechanism. To take into account the energy efficiency, since $\widetilde{\gamma_i}(\widetilde{x})$ is differentiable,  we can include it as a regularization term, training our network on a weighted combination of the costs:

\begin{equation}
\label{eq:gammacost}
\tilde{\gamma}_i(\widetilde{x}) = c_i\gamma_i(\widetilde{x}) +(1-c_i)\tilde{\gamma}_{i+1}(\widetilde{x})
\end{equation}
where $B$ is the number of EECs and $\widetilde{\gamma}_{B+1}(\widetilde{x})$ is the computational cost of exiting at the final exit of the EENN $\widetilde{x}$. For simplicity, we refer to $c_i$ as the confidence output at the $i^{th}$ EEC for a generic sample.

To compute the regularization, we adopt $\widetilde{\gamma}_{B+1}(\widetilde{x})$ instead of $F_M(\widetilde{x})$ since the ratio of exiting samples from a branch can be computed only at inference time. In the loss $L_{cost}$, the local confidences $c_i$ act as a proxy of these ratios.

The loss function $L_{cost}$ is implemented as follows:
\begin{equation}
\label{eq:costloss}
L_{cost}= \frac{\max(0,\widetilde{\gamma}_{B+1}(\widetilde{x})-\overline{F}_M)}{(F_M(x)-\overline{F}_M)}
\end{equation}
where $F_M(x)$ is the number of MACs computed by the backbone $x$ (i.e., no EECs) to process the input and $\overline{F}_M$ is the constraint enforced on $F_M(\widetilde{x})$. 

In the right term of Equation \eqref{eq:costloss}, the numerator relies on a $\max$ operator to apply the penalization in case of a violation of the constraint and the denominator computes the maximum possible violation since $F_M(x)$ is equal to $F_M(\widetilde{x})$ in the worst case, i.e., when all samples exit from the last EEC. In this perspective, $L_{cost}$ can be seen as the ratio of the constraint violation.

Second, we explain the regularization introduced to force the confidences to follow the operating point of the EENN. We aim at improving the accuracy of the EENN by tuning the confidences while considering the performance on a representative set. First, part of the original training set, containing an equal number of samples for each class, is extracted. We call this extracted set as \textit{Support Set} and the number of classes as $C$. The \textit{Support Set} is used to compute a matrix $C \times B$ named \textit{Support Matrix}. The \textit{Support Matrix} contains the average accuracies per class computed for each epoch on top of the support set. For each class, the average accuracy is the ratio of the corrected predicted samples in the set of samples of a given class in the support set. Once defined the \textit{Support Matrix}, we use a regression loss as a regularization term to force the local certainties (calculated on the batch samples) to follow the global accuracies (calculated on the \textit{Support Set}). Basically, we compute the Mean Squared Error (MSE) on two vectors of $B$ values computed per each sample as follows:
\begin{equation}
\label{eq:suploss}
L_{peak}=\text{MSE}(c_{local},SM_j)
\end{equation}
where the first vector, $c_{local}$ is the vector of local certanties and the second vector, $SM_{j}$ is the j-th row of the \textit{Support Matrix} corresponding to the label of the class of that sample. We initially train the EENN without the $L_{peak}$ term to train all the EECs and then apply this loss after $\mu_1 + \mu_2$ epochs when the EECs are working properly.

We now detail how the inference of the EENNs and the automatic setting of the thresholds are performed in NACHOS.

The inference is performed as in the scheme shown in Equation \eqref{eq:confidencethreshold}. If the confidence threshold in the prediction is met, a prediction is computed and the sample is not forward propagated. The confidence measure is computed as the Cumulative in the prediction~\cite{pomponieenn}.  

The thresholds are automatically set according to the constraints regarding the accuracy and the number of MACs. This is done by at first selecting the configuration of thresholds leading to the best accuracy in inference through a grid search. Then, if possible, starting from the second last exit the thresholds are reduced up to when the number of MACs is less than the constraint value while respecting the constraint regarding the accuracy. Note that the lower the thresholds, the higher the number of samples exiting from that EEC, hence reducing the overall number of MACs.

\subsection{MACs' formulation in the Search Strategy module}
\label{subsct:cmacs}

In this section, we describe the novel figure of merit named $F_{CM}(\widetilde{x})$ that endows $F_M(\widetilde{x})$ by handling the $\overline{F}_M$ constraint. This objective is optimized by the Search Strategy module. We define $\phi$ as the ratio of the admissible solutions in the population during the evolutionary process in the Search Strategy. The $F_{CM}(\widetilde{x})$ objective can be formulated as follows:
\begin{equation}
\label{eq:cmacs}
F_{CM}(\widetilde{x}) = \phi F_M(\widetilde{x}) + (1-\phi) \max(0,F_M(\widetilde{x}) - \overline{F}_M)
\end{equation}
The $\max$ operator eventually applies a penalty as in Equation \eqref{eq:costloss}. $\phi$ can be seen as an adaptive penalty term changing during the execution of the genetic algorithm.
Tipically $\phi$ is $\cong 0$ at the early stages of the NAS when there are a lot of unfeasible solutions and is $\cong 1$ at the final stages when there are more feasible solutions. The rationale is that $F_{CM}(\widetilde{x})$ aims at applying a stronger penalization to rapidly eliminate unfeasible solutions at the early stages and focus on optimizing the $F_M(\widetilde{x})$ at the final stages.

\section{Experimental results} 
\label{sct:experiments}
This section shows the results of the experimental campaign
aiming at assessing the effectiveness of NACHOS. In more
detail, the purpose of this section is to point out that NACHOS
can design effective network architectures
(i.e., providing high accuracy for the considered task), competitive with respect to state-of-the-art EENNs that have been manually designed by experts, while satisfying a constraint on the number of Multiply and Accumulate (MAC) operations performed by the EENNs at inference time.

\subsection{Dataset}
The datasets used in the experimental comparisons are CIFAR-10~\cite{cifar}, CINIC-10~\cite{cinic10}, SVHN~\cite{svhn}, and Imagenette~\cite{howard2020fastai}, a subset of 10 easily classified classes from Imagenet~\cite{imagenet}, whose number of classes is 1000 and image resolution is 224. The first three datasets consist of $32 \times 32$ color images while the resolution of Imagenette is $160 \times 160$, and represent 10 classes of objects. Regarding CIFAR-10, the training set comprises 60000 images, while the testing set comprises 10000 images.
Regarding CINIC-10, both the training set and the test set comprise 90000 images. Regarding SVHN, the training set comprises 73257 images, while the testing set comprises 26032 images. Regarding Imagenette, the training set comprises 9469 images while the testing set consists of 3925 images.

\begin{table*}[h!]
\centering
\caption{Results for the experimental campaign of NACHOS on all the datasets. Column labels explained in Table~\ref{tab:terminology}.}
\label{tab:NACHOS_complete}
\scalebox{0.75}{
\begin{tabular}{c|c|c|c|c|c|c|c|c|c}
\toprule
Dataset & Model & NAS & B & $F_M(\widetilde{x})$ (M) & $F_M(x)$ (M) & $F_A(\widetilde{x})$ (\%) & $F_A(x)$ (\%) & $\epsilon$ & Confidence \\
\midrule

\multirow{4}{*}{\rotatebox[origin=c]{0}{\small SVHN}} 
& \textbf{NACHOS} & yes & $3.00 \pm 0.71$ & $\bm{1.46 \pm 0.03}$ & $4.46 \pm 0.18$ & $79.96 \pm 0.84$ & $88.62 \pm 0.23$ & [$0.26 \pm 0.05$, $0.13 \pm 0.10$, $0.10$, –] & Cumulative \\ 
& EDANAS & yes & $4.60 \pm 0.55$ & $\underline{1.47 \pm 0.02}$ & $5.95 \pm 0.05$ & $77.98 \pm 0.24$ & $86.08 \pm 0.19$ & [$0.20 \pm 0.00$, $0.14 \pm 0.05$, $0.10 \pm 0.00$, $0.10 \pm 0.00$] & Score margin \\ 
& EEAlexnet & no & 5.00 & 263.51 & 464.48 & $\bm{89.00}$ & 93.10 & [0.70, 0.60, 0.60, 0.60] & Cumulative \\
& EEResnet20 & no & 10.00 & 70.20 & 315.26 & $\underline{88.53}$ & 93.28 & [0.70, 0.60, 0.60, 0.60, 0.60, 0.60, 0.60, 0.60, 0.60] & Cumulative \\
\midrule

\multirow{5}{*}{\rotatebox[origin=c]{0}{\small CIFAR-10}} 
& \textbf{NACHOS} & yes & $2.80 \pm 0.84$ & $\bm{2.44 \pm 0.02}$ & $5.78 \pm 0.19$ & $72.65 \pm 0.49$ & $71.68 \pm 0.31$ & [$0.20 \pm 0.00$, $0.24 \pm 0.05$, $0.10 \pm 0.00$, –] & Cumulative \\ 
& EDANAS & yes & $4.60 \pm 0.55$ & $\underline{2.47 \pm 0.01}$ & $6.00 \pm 0.04$ & $67.78 \pm 0.12$ & $67.48 \pm 0.23$ & [$0.20 \pm 0.00$, $0.18 \pm 0.04$, $0.12 \pm 0.04$, $0.10 \pm 0.00$] & Score margin \\ 
& EECNN & no & 2.00 & 7.48 & 25.14 & $\bm{76.50}$ & 77.20 & [0.10] & Max softmax \\
& EEAlexnet & no & 5.00 & 233.46 & 464.48 & $\underline{73.03}$ & 83.85 & [0.70, 0.60, 0.60, 0.60] & Cumulative \\
& EEResnet20 & no & 10.00 & 21.25 & 315.26 & 69.71 & 79.91 & [0.7, 0.60, 0.60, 0.60, 0.60, 0.60, 0.60, 0.60, 0.60] & Cumulative \\
\midrule

\multirow{4}{*}{\rotatebox[origin=c]{0}{\small CINIC-10}} 
& \textbf{NACHOS} & yes & $2.80 \pm 0.84$ & $2.42 \pm 0.01$ & $5.87 \pm 0.10$ & $60.85 \pm 0.30$ & $59.65 \pm 0.11$ & [$0.20 \pm 0.04$, $0.22 \pm 0.04$, $0.10 \pm 0.00$, –] & Cumulative \\ 
& EDANAS & yes & $4.20 \pm 0.45$ & $2.45 \pm 0.03$ & $6.03 \pm 0.02$ & $60.65 \pm 0.16$ & $59.67 \pm 0.17$ & [$0.20 \pm 0.00$, $0.18 \pm 0.04$, $0.10 \pm 0.00$, $0.10 \pm 0.00$] & Score margin \\ 
& EEAlexnet & no & 5.00 & 300.20 & 464.48 & $\bm{64.35}$ & 73.30 & [0.80, 0.80, 0.80, 0.80] & Cumulative \\
& EEResnet20 & no & 10.00 & 142.57 & 315.26 & $\underline{61.69}$ & 69.01 & [0.70, 0.60, 0.60, 0.60, 0.60, 0.60, 0.60, 0.60, 0.60] & Cumulative \\
\midrule

\multirow{3}{*}{\rotatebox[origin=c]{0}{\small Imagenette}} 
& \textbf{NACHOS} & yes & $3.00 \pm 0.71$ & $\underline{92.34 \pm 0.58}$ & $104.09 \pm 1.09$ & $\underline{94.89 \pm 0.15}$ & $95.53 \pm 0.08$ & [–, $0.63 \pm 0.06$, $0.68 \pm 0.04$, $0.65 \pm 0.07$] & Cumulative \\
& EDANAS & yes & $4.40 \pm 0.55$ & $\bm{81.92 \pm 1.11}$ & $231.34 \pm 4.49$ & $90.76 \pm 0.54$ & $96.53 \pm 0.38$ & [$0.10\pm0.00$, $0.20\pm0.00$, $0.14\pm0.05$, $0.15\pm0.07$] & Score margin \\
& EEEfficientNet-B0 & no & 4.00 & 182.77 & 201.13 & $\bm{95.14}$ & 97.25 & [0.50, 0.50, 0.50] & Cumulative \\
\bottomrule
\end{tabular}
}
\end{table*}

\begin{table*}[h!]
\centering
\caption{Computational costs ($\gamma$) and utilization (U) for the experimental campaign of NACHOS on the four datasets.}
\label{tab:gammacosts}
\scalebox{0.75}{
\begin{tabular}{@{}c|c|c|c@{}}
\toprule
Dataset & Model & Computational Costs $\gamma$ (M) & Utilization U (\%) \\ \midrule

\multirow{5}{*}{SVHN}
& \textbf{NACHOS} & [$1.12 \pm 0.08$, $1.85 \pm 0.11$, $2.5 \pm 0.00$, –, $4.46 \pm 0.18$] & [$58.78 \pm 21.89$, $39.91 \pm 24.51$, $0.00 \pm 0.00$, –, $1.31 \pm 2.92$] \\
& EDANAS & [$1.13 \pm 0.12$, $1.51 \pm 0.06$, $2.07 \pm 0.04$, $4.01 \pm 0.03$, $5.95 \pm 0.05$] & [$24.97 \pm 16.69$, $82.94 \pm 20.04$, $1.28 \pm 2.64$, $0.24 \pm 0.43$, $0.56 \pm 0.55$] \\
& EEAlexnet & [87.27, 201.59, 380.82, 440.54, 464.48] & [23.00, 34.00, 27.00, 10.00, 6.00] \\
& EEResnet20 & [19.57, 24.73, 29.89, 35.06, 43.99, 64.63, 85.28, 158.97, 239.48, 315.26] & [0.00, 0.00, 0.00, 34.00, 13.00, 34.00, 11.00, 2.00, 2.00, 4.00] \\
& EEEfficientNet-B0 & – & – \\
\midrule

\multirow{5}{*}{CIFAR-10}
& \textbf{NACHOS} & [$1.60 \pm 0.0$, $2.18 \pm 0.31$, $3.40 \pm 0.44$, –, $5.78 \pm 0.19$] & [$41.31 \pm 30.37$, $74.31 \pm 30.32$, $0.00 \pm 0.00$, –, $0.90 \pm 1.52$] \\
& EDANAS & [$1.11 \pm 0.02$, $1.49 \pm 0.04$, $2.03 \pm 0.02$, $4.03 \pm 0.03$, $6.00 \pm 0.04$] & [$0.87 \pm 1.17$, $24.58 \pm 7.56$, $51.06 \pm 12.65$, $18.78 \pm 10.17$, $5.08 \pm 4.65$] \\
& EEAlexnet & [87.27, 201.59, 380.82, 440.54, 464.48] & [38.00, 27.00, 19.00, 4.00, 12.00] \\
& EEResnet20 & [19.57, 24.73, 29.89, 35.06, 43.99, 64.63, 85.28, 158.97, 239.48, 315.26] & [6.00, 0.00, 0.00, 33.00, 16.00, 18.00, 8.00, 9.00, 5.80, 4.20] \\
& EEEfficientNet-B0 & – & - \\
\midrule

\multirow{5}{*}{CINIC-10}
& \textbf{NACHOS} & [$1.80 \pm 0.20$, $2.45 \pm 0.50$, $3.70 \pm 0.42$, –, $5.87 \pm 0.10$] & [$69.75 \pm 26.03$, $30.25 \pm 26.03$, $0.00 \pm 0.00$, –, –] \\
& EDANAS & [$1.12 \pm 0.00$, $1.48 \pm 0.04$, $2.09 \pm 0.03$, $4.06 \pm 0.05$, $6.03 \pm 0.02$] & [$3.30 \pm 0.00$, $25.74 \pm 9.86$, $52.38 \pm 6.46$, $15.8 \pm 6.79$, $5.42 \pm 3.48$] \\
& EEAlexnet & [87.27, 201.59, 380.82, 440.54, 464.48] & [16.00, 34.00, 13.00, 20.00, 17.00] \\
& EEResnet20 & [19.57, 24.73, 29.89, 35.06, 43.99, 64.63, 85.28, 158.97, 239.48, 315.26] & [0.00, 0.00, 0.00, 0.00, 14.00, 24.00, 20.00, 0.40, 28.00, 10.00] \\
& EEEfficientNet-B0 & – & – \\
\midrule

\multirow{5}{*}{Imagenette}
& \textbf{NACHOS} & [–, $46.00 \pm 5.29$, $59.81 \pm 3.63$, $70.21 \pm 6.78$, $104.09 \pm 1.09$] & [$4.67 \pm 3.33$, $14.31 \pm 9.70$, $26.02 \pm 14.98$, –, $72.42 \pm 7.22$] \\
& EDANAS & [$59.58 \pm 1.90$, $79.01 \pm 1.83$, $92.86 \pm 1.80$, $149.77 \pm 4.38$, $231.34 \pm 4.49$] & [$0.37 \pm 0.35$, $96.76 \pm 1.86$, $0.70 \pm 0.47$, $2.10 \pm 0.42$, $1.48 \pm 0.91$] \\
& EEAlexnet & – & – \\
& EEResnet20 & – & – \\
& EEEfficientNet-B0 & [50.74, 103.16, 140.96, 201.13] & [0.00, 6.34, 20.19, 73.47] \\
\bottomrule
\end{tabular}
}
\end{table*}

\begin{table}[h!]
\centering
\caption{ECE scores for the experimental campaign of NACHOS on CIFAR-10 dataset.
}
\label{tab:eceCIFAR-10}
\scalebox{0.8}{
\begin{tabular}{@{}c|c@{}}
\toprule
Model  &  ECE scores (\%) \\ \midrule
\textbf{NACHOS} & [$18.23\pm0.00$, $15.70\pm0.22$, $4.53\pm0.15$, -, $1.60\pm0.11$] \\ 
\Xhline{2\arrayrulewidth}
%\textbf{NACHOS\_NP}  & [14.20, 2.81] \\ 
%\Xhline{2\arrayrulewidth}
EEAlexnet  & [13.79, 40.55, 59.07, 71.44, 2.26] \\ 
\midrule
EEResnet20 & [16.51, 89.80, 77.80, 28.39, 62.35, 58.10, 46.60, 66.65, 62.25, 2.30] \\ 
\midrule
\end{tabular}
}
\end{table}

\subsection{Experimental setup} 
% First, we report the most important terms used to describe the NACHOS framework in Table~\ref{tab:terminology}.

\begin{figure}
\begin{minipage}[c]{\linewidth}
\centering
\includegraphics[width=0.9\linewidth]{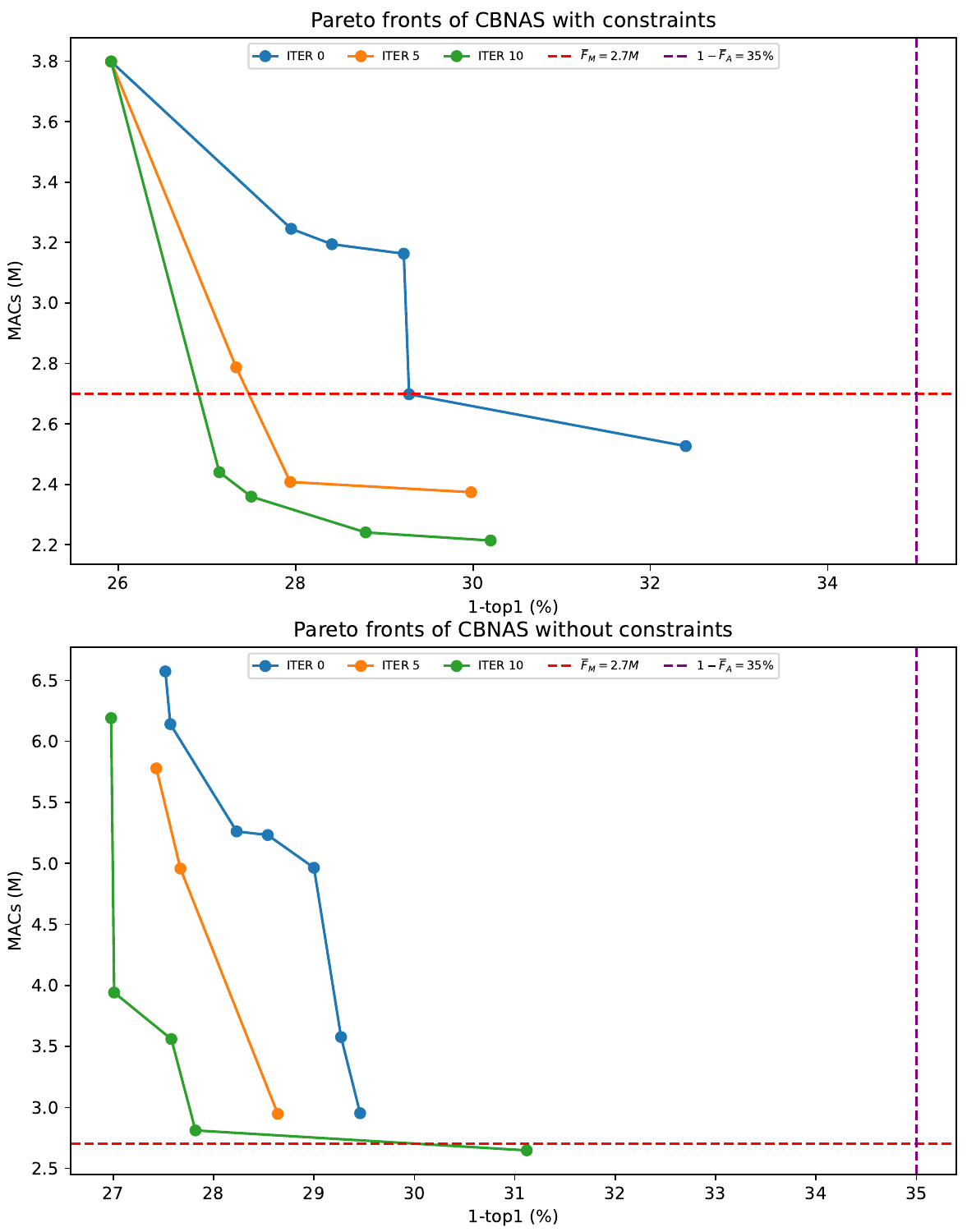}
\caption{Comparison between the Pareto fronts of runs of NACHOS enforcing and not enforcing constraints in the search on CIFAR-10 respectively, while setting the same seed.}
\label{fig:ComparisonParetoFronts}
\end{minipage}

\vfill
\begin{minipage}[c]{\linewidth}
\centering
\includegraphics[width=0.8\linewidth]{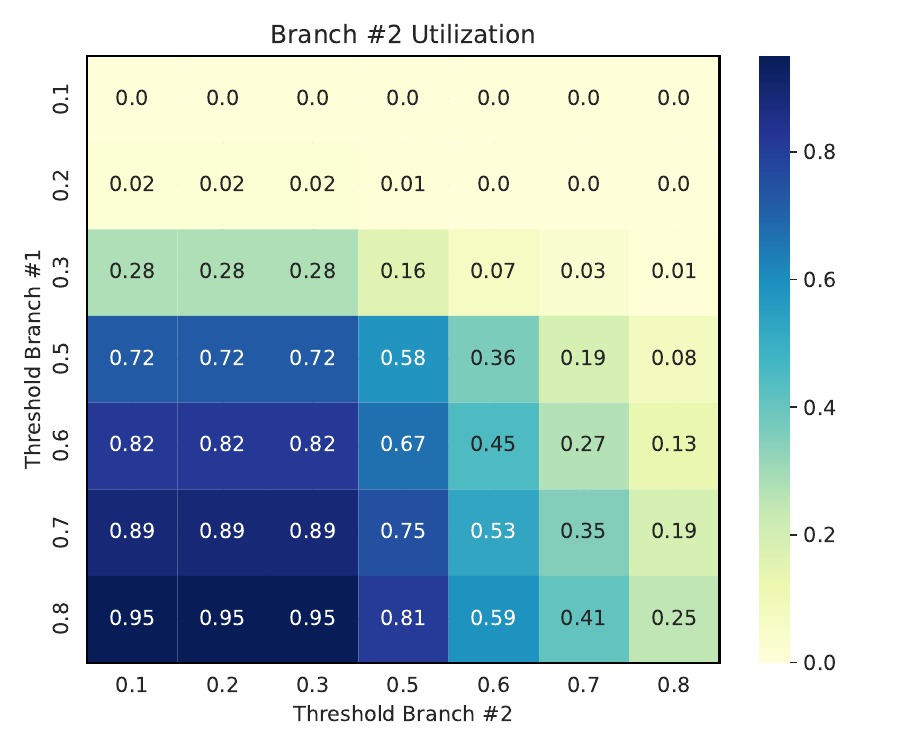}
\caption{Heatmap of the utilization of the second EEC of a model with three exits (B=3) found by NACHOS\_NC varying the thresholds of the first two EECs.}
\label{fig:heatmapsEENN}
\end{minipage}
\end{figure}

The setup of the experiments is the following:
\begin{itemize}
\item the 1st objective is $F_A(\widetilde{x})$ and the 2nd objective is $F_M(\widetilde{x})$;
\item the training epochs are set as $\{\mu_1=10, \mu_2=5, \mu_3=5\}$;
\item the weights of the training loss function shown in Equation \eqref{eq:trainlossNACHOS} are set as $\{\omega_1=1.0, \omega_2=1.0, \omega_3=1.0\}$ and $\omega_3=0.0$ in the step following the fine-tuning (i.e., no $L\_{peak}$ is applied);
\item the initial number of samples populating the archive $N_{start}$ is
set to 100;
\item the maximum number of iterations $N_{iter}$ is 30 and the number of epochs to trigger the convergence criterion $N_{tol}$ is 10;
\item considering our constrained mobile setting, we chose the pretrained MobileNetV3-based supernet, being the lightest in terms of MACs. We highlight that our framework is flexible enough to be applied to other CNN-based types of supernets. The hyperparameters of $\Omega_x$ refer to the depth, the kernel size, and the expansion rate (the architecture search space and building blocks are detailed in \cite{mobilenetv3}) while the resolution is fixed at 32 which is the size of the images of the datasets used in the experimental results;
\item the number of exit points (i.e., the points of the network
where EECs are possibly placed) is set to 4 since it is a reasonable arbitrary number of EECs in an EENN and is also chosen by EDANAS;
\item the constraints are set as $\overline{F}_A=65\%$, $\overline{F}_M=2.7M$ on CIFAR-10; $\overline{F}_A=50\%,$ and $\overline{F}_M=2.7M$ on CINIC-10;
$\overline{F}_A=75\%,$ and $\overline{F}_M=1.5M$ on SVHN; $\overline{F}_A=90\%,$ and $\overline{F}_M=93M$ on Imagenette. The constraints are chosen by considering the results of EDANAS on the different datasets to test whether the searched models perform better in a similar setting.
\item the computational resources required: search time of 2 GPU days on a single GPU (NVIDIA A40) for both NACHOS and EDANAS. Their search times are the same since NACHOS takes more training time in the evaluation of the NNs but converges faster in terms of NAS epochs thanks to constraints as shown in Fig. \ref{fig:ComparisonParetoFronts}.
\item the results of all the tested NAS methods, NACHOS and EDANAS, reports the mean and the standard deviation obtained by aggregating the results of the search over 5 different seeds following the best practice for NAS evaluations suggested in \cite{bestpractices}. We ometted the mean and standard deviation of $\gamma$, $\epsilon$, and $U$ when the corresponding early exit has never been selected among the runs and placed a '-' instead.
\end{itemize}

%\changemarkertable{
%All in all, this setting ensures the correct functioning of the NAS, i.e., it ensures that the evaluation of the EENNs provides enough, and consistent data to train the surrogate models and make them produce reliable predictions. Their reliability is measured according to a metric named Kendall's Tau \cite{kendalltau}.}

Additionally, we report the hyperparameters of the genetic algorithm NSGA-II:
\begin{itemize}
\item population size: 40;
\item initial population: non-dominated solutions of the previous NAS iteration;
\item crossover operator: two-point crossover with crossover probability=0.9;
\item mutation operator: polynomial mutation with distribution index=1.0;
\item selection operator: tournament selection with crowding distance sorting;
\item number of generations: 20;
\item survival: only the eight best solutions are kept according to ranking (Pareto dominance) and crowding distance sorting.
\end{itemize}
This setup allows effective convergence toward optimal trade-offs while preserving the diversity of the solutions, reducing the risk of premature convergence.

$N_{start}, N_{iter}$, the hyperparameters of the genetic algorithm, and the training parameters such as batch size and learning rate follow the experimental setup of MSuNAS and EDANAS.

The output set of optimal architectures $X^{\circ}$ of each search contains one EENN. This architecture is Pareto Optimal using the best
trade-off between $F_A(\widetilde{x})$ and $F_M(\widetilde{x})$ as the selection
criterion. We refer to these networks as NACHOS.

% \begin{figure}[t!]%
% 	\centering
% 	\includegraphics[width=0.9\linewidth]{Images/PF_Comparison.pdf}
% \caption{Comparison between the Pareto fronts of NACHOS enforcing and not enforcing constraints in the search on CIFAR-10 respectively.}
% \label{fig:ComparisonParetoFronts}
% \end{figure}

% \begin{figure}[t!]%
% 	\centering
% 	\includegraphics[width=0.8\linewidth]{Images/Branch2Utilization.pdf}
% \caption{Heatmap of the utilization of the second EEC of the NACHOS\_NC varying the thresholds of the first two EECs.}
% \label{fig:heatmapsEENN}
% \end{figure}

As a comparison, we considered the following state-of-the-art EENNs.
\begin{enumerate}
\item the EENN found by NACHOS without using the $L_{cost}$ regularization by selecting the best trade-off. We refer to this network as NACHOS\_NC (NACHOS No Constraints).
\item the EENN found by NACHOS without using the $L_{peak}$ regularization by selecting the best trade-off. We refer to this network as NACHOS\_NP (NACHOS No Peak).
\item the EENN found by EDANAS by selecting the best trade-off. We refer to these networks as $EDANAS$.
\item an EENN~\cite{disabato_roveri_2018} with an auxiliary classifier consisting of two fully connected layers with 1024 and 10 neurons placed after the first max-pooling. We refer to this network as EECNN;
\item AlexNet, and ResNet-20~\cite{resnet}  modified to handle $32 \times 32$ images with multiple EECs~\cite{pomponieenn}.
An EEC is placed after each layer for AlexNet, resulting in 5 EECs (including the last one), while the
branches in ResNet20 are placed after each block, resulting in a total of 10 branches. Each
branch is composed of a convolutional layer of 128 filters, followed by a max pooling
operation, if the dimensions of the image are big enough. This block is followed by a
ReLU activation function and a classification linear layer. We refer to these networks as EEAlexNet and EEResnet20, respectively.
\item EfficientNet-B0~\cite{efficientnet} with 3 EECs placed by the computational-aware procedure explained in Algorithm~\ref{alg:ofaadapter}. We refer to this network as EEEfficientNet-B0.
\begin{comment}
\item AlexNet, VGG-13, and ResNet-18 modified to handle 32x32 images with multiple EECs~\cite{scardapane_differentiable_2020}. The networks are enriched with 5 EECs for AlexNet and 9 for the other two architectures. Each EEC consists of 
a convolutional layer with 128 filters, a max-pooling, and a fully-connected layer. We refer to these networks as \textit{EEAlexNet}, \textit{EEVGG-13}, and \textit{EEResNet-18};
\end{comment}
\end{enumerate}

EDANAS has been run with the same configuration shown in ~\cite{edanas} but differently, the resolution of the networks has been fixed to $160 \times 160$ on Imagenette and $32 \times 32$ on the other datasets.

The expert-designed networks have been trained with the following methodologies. EECNN has been trained by joint training. EEAlexNet, EEResnet20, and EEEfficientNet-B0 have been trained by the same differentiable training adopted in NACHOS presented in the work~\cite{pomponieenn}. In particular, the backbone is initially trained from scratch (i.e., no fine-tuning) for $\mu_1$ epochs, and then the standalone $L_{acc}$ loss function is optimized for $\mu_2+\mu_3$ epochs during the training of the whole EENN. The thresholds are chosen with the same grid search on the same intervals used in NACHOS but considering no constraints.

We explain briefly the choice of the benchmarks. EDANAS is the only other framework in the literature optimizing jointly the backbone and the EECs. EECNN is a valid example of a CNN with a good trade-off between number of MACs and top1 accuracy. EEAlexNet, EEResnet20, and EEEfficientNet-B0 trained for the same number of epochs of NACHOS (differently from EECNN), are taken to ensure fair benchmarking with state-of-the-art EENNs. Additionally, NACHOS is tested without $L_{peak}$ and $L_{cost}$ regularizations to verify their effectiveness in the framework.

\begin{comment}
The work \cite{bestpractices} reasonably advocates multiple runs to account for NAS search stochasticity when computationally feasible. In our case, stochasticity is reduced by using an OFA supernet with fixed pretrained weights \cite{ofa} and the robust NSGA-II genetic algorithm \cite{nsga-II}. Similar settings in MSuNAS \cite{msunas} and EDANAS \cite{edanas} relied on a single search run to report the performance of the NAS. Additionally, we enforced constraints to guide exploration toward the target search space. Given these factors, we performed only one search run for each experiment.
\end{comment}

\begin{table*}[t!]
\centering
\caption{Results for the ablation studies of NACHOS on all datasets. Column labels are explained in Table~\ref{tab:terminology}.}
\label{tab:ablNACHOS_complete}
\scalebox{0.75}{
\begin{tabular}{c|c|c|c|c|c|c|c|c|c}
\toprule
Dataset & Model & NAS & B & $F_M(\widetilde{x})$ (M) & $F_M(x)$ (M) & $F_A(\widetilde{x})$ (\%) & $F_A(x)$ (\%) & $\epsilon$ & Confidence \\
\midrule

\multirow{3}{*}{SVHN} 
& \textbf{NACHOS} & yes & $3.00 \pm 0.71$ & \underline{$1.46 \pm 0.03$} & $4.46 \pm 0.18$ & \underline{$79.96 \pm 0.84$} & $88.62 \pm 0.23$ & [$0.26 \pm 0.05$, $0.13 \pm 0.10$, $0.10$, –] & Cumulative \\ 
& \textbf{NACHOS\_NC} & yes & $2.80 \pm 0.84$ & $1.95 \pm 0.10$ & $4.74 \pm 0.08$ & $\bm{84.59 \pm 0.47}$ & $88.24 \pm 0.27$ & [$0.67 \pm 0.06$, $0.16 \pm 0.05$, $0.10 \pm 0.00$, –] & Cumulative \\ 
& \textbf{NACHOS\_NP} & yes & $2.80 \pm 0.84$ & $\bm{1.45 \pm 0.02}$ & $4.62 \pm 0.18$ & $76.94 \pm 0.63$ & $89.04 \pm 0.43$ & [$0.18 \pm 0.08$, $0.07 \pm 0.11$, $0.10 \pm 0.00$, –] & Cumulative \\
\Xhline{2\arrayrulewidth}

\multirow{3}{*}{CIFAR-10} 
& \textbf{NACHOS} & yes & $2.80 \pm 0.84$ & \underline{$2.44 \pm 0.02$} & $5.78 \pm 0.19$ & $\bm{72.65 \pm 0.49}$ & $71.68 \pm 0.31$ & [$0.20 \pm 0.00$, $0.24 \pm 0.05$, $0.10 \pm 0.00$, –] & Cumulative \\ 
& \textbf{NACHOS\_NC} & yes & $3.00 \pm 0.70$ & $2.91 \pm 0.11$ & $6.43 \pm 0.12$ & \underline{$72.25 \pm 0.22$} & $71.38 \pm 0.33$ & [$0.65 \pm 0.06$, $0.16 \pm 0.05$] & Cumulative \\ 
& \textbf{NACHOS\_NP} & yes & $2.40 \pm 0.55$ & $\bm{2.40 \pm 0.05}$ & $6.25 \pm 0.42$ & $71.38 \pm 0.47$ & $68.23 \pm 0.59$ & [–, $0.26 \pm 0.05$, $0.15 \pm 0.07$, –] & Cumulative \\
\Xhline{2\arrayrulewidth}

\multirow{3}{*}{CINIC-10} 
& \textbf{NACHOS} & yes & $2.80 \pm 0.84$ & \underline{$2.42 \pm 0.01$} & $5.87 \pm 0.10$ & $\bm{60.85 \pm 0.30}$ & $59.65 \pm 0.11$ & [$0.20 \pm 0.04$, $0.22 \pm 0.04$, $0.10 \pm 0.00$, –] & Cumulative \\ 
& \textbf{NACHOS\_NC} & yes & $3.00 \pm 0.70$ & $3.74 \pm 0.13$ & $6.85 \pm 0.11$ & $59.96 \pm 0.12$ & $59.51 \pm 0.29$ & [$0.70 \pm 0.00$, $0.60 \pm 0.17$, $0.58 \pm 0.10$, –] & Cumulative \\ 
& \textbf{NACHOS\_NP} & yes & $2.40 \pm 0.55$ & $\bm{2.41 \pm 0.06}$ & $4.79 \pm 0.12$ & \underline{$60.16 \pm 0.34$} & $59.63 \pm 0.16$ & [$0.20 \pm 0.00$, $0.24 \pm 0.05$, –, –] & Cumulative \\
\Xhline{2\arrayrulewidth}

\multirow{3}{*}{Imagenette} 
& \textbf{NACHOS} & yes & $3.00 \pm 0.71$ & \underline{$92.34 \pm 0.58$} & $104.09 \pm 1.09$ & \underline{$94.89 \pm 0.15$} & $95.53 \pm 0.08$ & [–, $0.63 \pm 0.06$, $0.68 \pm 0.04$, $0.65 \pm 0.07$] & Cumulative \\ 
& \textbf{NACHOS\_NC} & yes & $3.20 \pm 0.45$ & $\bm{118.01 \pm 1.47}$ & $121.30 \pm 3.15$ & $\bm{97.36 \pm 0.11}$ & $96.46 \pm 0.25$ & [–, $0.70 \pm 0.00$, $0.68 \pm 0.04$, $0.66 \pm 0.05$] & Cumulative \\ 
& \textbf{NACHOS\_NP} & yes & $2.40 \pm 0.55$ & $84.38 \pm 0.33$ & $98.88 \pm 0.41$ & $85.24 \pm 0.15$ & $96.65 \pm 0.25$ & [–, $0.15 \pm 0.06$, $0.17 \pm 0.06$, –] & Cumulative \\
\Xhline{2\arrayrulewidth}

\end{tabular}
}
\end{table*}

\begin{table*}[h!]
\centering
\caption{Computational costs ($\gamma$) and utilization (U) for the ablation studies of NACHOS on the four datasets.}
\label{tab:ablgammacosts}
\scalebox{0.75}{
\begin{tabular}{@{}c|c|c|c@{}}
\toprule
Dataset & Model & Computational Costs $\gamma$ (M) & Utilization U (\%) \\ \midrule

\multirow{3}{*}{SVHN} 
& \textbf{NACHOS} & [$1.12 \pm 0.08$, $1.85 \pm 0.11$, $2.53 \pm 0.00$, –, $4.46 \pm 0.18$] & [$58.78 \pm 21.89$, $39.91 \pm 24.51$, $0.00 \pm 0.00$, –, $1.31 \pm 2.92$] \\
& \textbf{NACHOS\_NC} & [$1.16 \pm 0.14$, $1.86 \pm 0.18$, $2.60 \pm 0.00$, –, $4.74 \pm 0.08$] & [$22.65 \pm 2.61$, $72.29 \pm 27.00$, $49.58 \pm 0.00$, –, $4.20 \pm 6.26$] \\
& \textbf{NACHOS\_NP} & [$1.20 \pm 0.07$, $2.03 \pm 0.07$, $2.81 \pm 0.00$, –, $4.62 \pm 0.18$] & [$78.29 \pm 12.81$, $30.34 \pm 6.92$, $0.00 \pm 0.00$, –, $3.50 \pm 4.84$] \\
\midrule

\multirow{3}{*}{CIFAR-10} 
& \textbf{NACHOS} & [$1.60 \pm 0.00$, $2.18 \pm 0.31$, $3.40 \pm 0.44$, –, $5.78 \pm 0.19$] & [$41.31 \pm 30.37$, $74.31 \pm 30.32$, $0.00 \pm 0.00$, –, $0.90 \pm 1.52$] \\
& \textbf{NACHOS\_NC} & [$1.68 \pm 0.06$, $2.57 \pm 0.38$, $3.67 \pm 0.24$, –, $6.43 \pm 0.12$] & [$44.54 \pm 22.45$, $42.02 \pm 32.35$, $17.07 \pm 2.23$, –, $15.41 \pm 8.97$] \\
& \textbf{NACHOS\_NP} & [–, $2.01 \pm 0.27$, $3.05 \pm 0.08$, –, $6.25 \pm 0.42$] & [$55.2 \pm 1.91$, $74.39 \pm 28.54$, –, –, $3.53 \pm 3.27$] \\
\midrule

\multirow{3}{*}{CINIC-10} 
& \textbf{NACHOS} & [$1.80 \pm 0.20$, $2.45 \pm 0.50$, $3.70 \pm 0.42$, –, $5.87 \pm 0.10$] & [$69.75 \pm 26.03$, $30.25 \pm 26.03$, $0.00 \pm 0.00$, –, –] \\
& \textbf{NACHOS\_NC} & [$1.50 \pm 0.00$, $3.03 \pm 0.29$, $3.77 \pm 0.12$, –, $6.85 \pm 0.11$] & [$9.87 \pm 0.00$, $39.70 \pm 27.63$, $74.61 \pm 17.79$, –, $6.58 \pm 11.93$] \\
& \textbf{NACHOS\_NP} & [$1.25 \pm 0.07$, $2.34 \pm 0.12$, $2.95 \pm 0.21$, –, $4.79 \pm 0.12$] & [$63.28 \pm 2.86$, $60.85 \pm 44.39$, $12.25 \pm 0.95$, –, $8.94 \pm 3.21$] \\
\midrule

\multirow{3}{*}{Imagenette} 
& \textbf{NACHOS} & [–, $46.00 \pm 5.29$, $59.81 \pm 3.63$, $70.21 \pm 6.78$, $104.09 \pm 1.09$] & [$4.67 \pm 3.33$, $14.31 \pm 9.70$, $26.02 \pm 14.98$, –, $72.42 \pm 7.22$] \\
& \textbf{NACHOS\_NC} & [–, $52.35 \pm 0.00$, $65.06 \pm 4.30$, $92.06 \pm 3.08$, $121.30 \pm 3.15$] & [–, $0.0 \pm 0.0$, $0.79 \pm 1.09$, $8.77 \pm 9.54$, $90.44 \pm 8.88$] \\
& \textbf{NACHOS\_NP} & [–, $51.00 \pm 2.95$, $62.67 \pm 2.52$, –, $98.88 \pm 0.41$] & [–, $24.55 \pm 8.93$, $24.16 \pm 12.65$, –, $65.86 \pm 3.22$] \\
\bottomrule
\end{tabular}
}
\end{table*}

\subsection{Analysis of the results} 
\label{subsct:results}
% \changemarkertable{
% The experimental results of NACHOS on the four different datasets are presented in Table~\ref{tab:NACHOSCIFAR-10}, Table~\ref{tab:NACHOScinic10}, and Table~\ref{tab:NACHOSsvhn}, and Table~\ref{tab:NACHOSImagenette}', respectively}
The experimental results of NACHOS are presented in Table \ref{tab:NACHOS_complete}. The column NAS specifies if the network is designed by a NAS procedure (yes) or not (no). The column B specifies the number of EECs (including the final exit). The column $F_M(\widetilde{x})$ shows the number of MACs in EE inference \footnote{The number of Floating Point Operations (FLOPS), another popular metric regarding inference time, can be approximately computed by the number of MACs as $ \text{FLOPS} \approx 2 \times \text{MACs}$}. The column $F_A(\widetilde{x})$ shows the top1 accuracy in EE inference. The column $F_A(x)$ shows the top1 accuracy in inference by using only the backbone of the EENN. The column $\epsilon$ shows the thresholds of the EECs (apart from the last classifier not having a thresholding). The column Confidence shows the technique used to estimate the confidence in the prediction. 

Additionally, the computational costs $\gamma_i$, i.e., the amounts of MACs to process
an input at the $i^{th}$ exit of an EENN, and the utilizations $U_i$, i.e., the exit ratios of the EECs, of the models presented in the aforementioned tables are reported in Table~\ref{tab:gammacosts}. We highlight that the costs of EEAlexnet and EEResnet20 are the same for SVHN, CIFAR-10, and CINIC-10 since they refer to the same model.

We highlight that the high utilizations of some EECs in the presented experiments are due to the strong constraint on the computational demands.

Comparing NACHOS with the other state-of-the-art EENNs we show that NACHOS is effective in selecting a competitive network in terms of best trade-off between the two objectives of the search. In particular, NACHOS is able to find an EENN with similar computational demand to EDANAS with similar accuracy on CINIC-10 while improving the accuracy on CIFAR-10, SVHN, and Imagenette. Moreover, NACHOS is competitive with the expert-designed models on CIFAR-10, CINIC-10, and Imagenette while it performs worse on SVHN. We highlight that the computational demand of the expert-designed models is significantly higher with respect to NACHOS.

Additionally, the Expected Calibration Error (ECE) scores for the experimental campaign of NACHOS on CIFAR-10 dataset are presented in Table~\ref{tab:eceCIFAR-10}. ECE is a popular metric to measure the calibration of the EECs \cite{naeini2015obtaining}. The calibration is an important aspect of EENNs related to their capability of estimating properly the confidences on input samples.

Table~\ref{tab:eceCIFAR-10} shows that the ECE scores of the NAS model are good pointing out that well-calibrated EENNs are found by the NACHOS method. However, we expect calibration to play a major role in NACHOS when considering networks with higher computational demands since, for instance, deeper networks tend to be more miscalibrated~\cite{guo_calibration_2017} (e.g., with Resnet-based OFA). This aspect will be addressed in future works of the paper.

\subsection{Ablation studies}
\label{subsct:ablation}

\begin{figure*}[h!]
\begingroup
%\scalebox{0.9}{
    \centering
\subfigure[Distribution of confidences in NACHOS model.]{
        \includegraphics[width=0.5\textwidth]{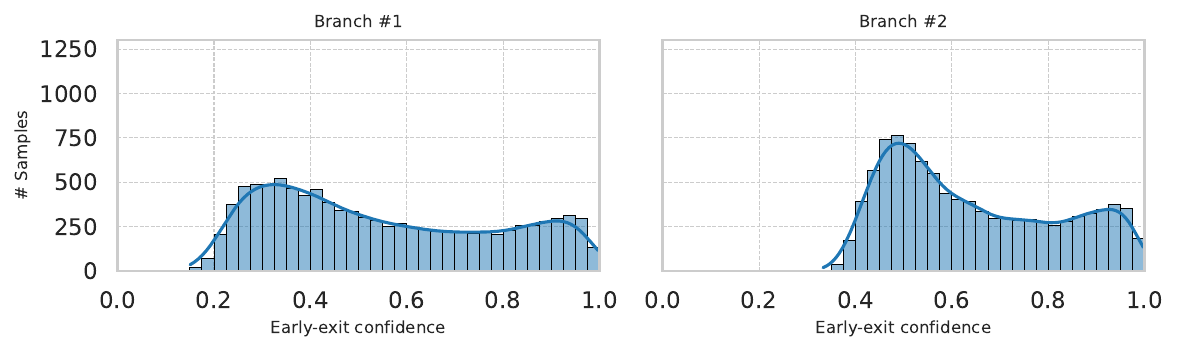}
        \label{fig:distconf1}
    }
\subfigure[Distribution of confidences in NACHOS\_NP model.]{
        \includegraphics[width=0.5\textwidth]{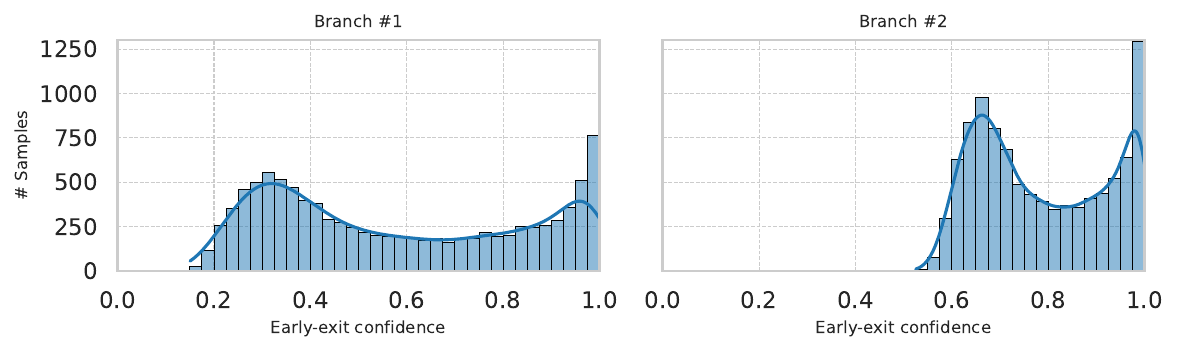}
        \label{fig:distconf2}
    }
    \subfigure[Distribution of confidences in NACHOS\_NC model.]{
        \includegraphics[width=0.9\textwidth]{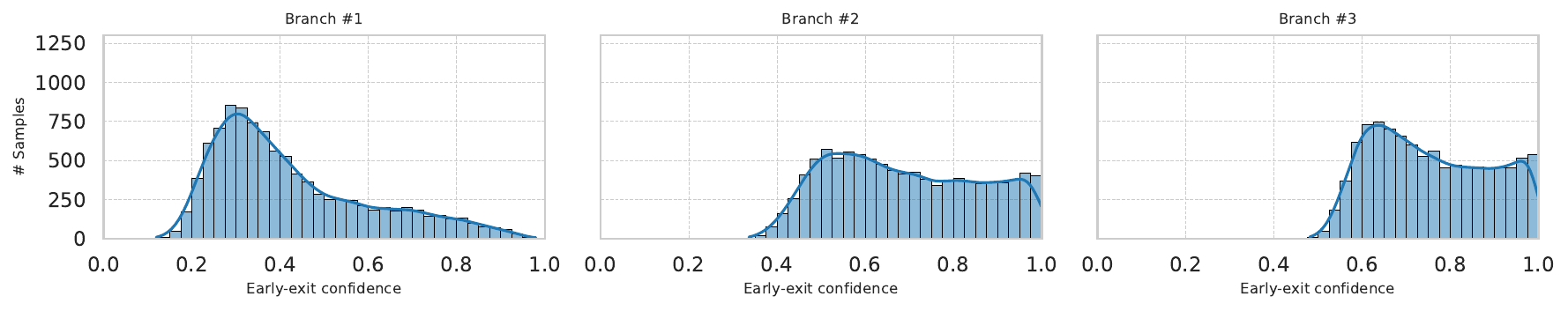}
        \label{fig:distconf3}
    }
 %   }
   \endgroup \caption{Comparison between the distribution of confidences of NACHOS models trained by applying different regularization while setting the same seed.}
    \label{fig:distconfs}
\end{figure*}

This section tests the effectiveness of the regularizations $L_{cost}$ and $L_{peak}$ introduced in Section~\ref{subsct:training}.

The results of the models NACHOS, NACHOS\_NC and NACHOS\_NP found on the four datasets are reported in Table \ref{tab:ablNACHOS_complete}. Additionally, the computational costs and the utilizations of the EECs of the models NACHOS, NACHOS\_NC, and NACHOS\_NP found on the four datasets are reported in Table \ref{tab:ablgammacosts}. 
% Table~\ref{tab:ablNACHOSCIFAR-10}, Table~\ref{tab:ablNACHOScinic10}, and Table~\ref{tab:ablNACHOSsvhn}. 

In the NACHOS experiments, comparing NACHOS with NACHOS\_NC,  we aim at testing the effectiveness of the $L_{cost}$ regularization in selecting an EENN satisfying the given constraints. The results show that the constraints are satisfied in NACHOS while are not satisfied in NACHOS\_NC, 
Additionally, the model found shows even better accuracy on CIFAR-10 and SVHN.
Additionally, Fig.~\ref{fig:ComparisonParetoFronts} points out that NACHOS by using constraints and the $L_{cost}$ regularization is able to reach faster the region of solutions satisfying the constraints, delimited by the red horizontal line and the purple vertical line in the figure, with respect to the unconstrained counterpart. Finally, we can argue that a NAS by enforcing constraints is able to find an admissible solution with a significant lower number of iterations. Indeed, NACHOS and EDANAS models are found by NACHOS in 10 iterations and by EDANAS in 30 iterations, respectively.
% 
\begin{comment}
\begin{figure}[t]%
	\centering
	\includegraphics[width=0.9\linewidth]{Images/PF_Comparison.pdf}
\caption{Comparison between the Pareto fronts of NACHOS enforcing and not enforcing constraints in the search on CIFAR-10 respectively.}
\label{fig:ComparisonParetoFronts}
\end{figure}
\end{comment}
% 
Fig.~\ref{fig:heatmapsEENN} shows the utilization of the second EEC of a model with three exits (B=3) found by NACHOS\_NC on CIFAR-10 varying the thresholds of the first two EECS. This visualizes how the thresholds affect the behaviour of the Cumulative method in triggering the early exit in the EECs.
% 
\begin{comment}
\begin{figure}[t!]%
	\centering
	\includegraphics[width=0.8\linewidth]{Images/Branch2Utilization.pdf}
\caption{Heatmap of the utilization of the second EEC of the NACHOS\_NC varying the thresholds of the first two EECs.}
\label{fig:heatmapsEENN}
\end{figure}
\end{comment}
% 
In the NACHOS experiments, comparing NACHOS with NACHOS\_NP we aim at testing the effectiveness of the $L_{peak}$ regularization in improving the accuracy of the EENN. An increase in accuracy is seen on CIFAR-10, SVHN, and Imagenette while the accuracy is approximately the same on CINIC-10 and maintains a similar computational complexity. 

Lastly, we report in Fig.~\ref{fig:distconfs} the distributions of the confidences in NACHOS, NACHOS\_NP, and NACHOS\_NC models obtained with the same seed. We can point out that the $L_{peak}$ is effective in aligning the confidences towards the operating point of the EECs. This is evident in the decrease of the number of samples for very high confidences for both EECs and the corresponding increase of samples towards the operating point in the NACHOS model with respect to the NACHOS\_NP model. Moreover, the $L_{cost}$ is effective in favoring the training of the early fraction of the EENN. This is seen in the higher number of samples for the higher confidences in the NACHOS model in the first EEC with respect to the NACHOS\_NC model.

\subsection{Statistical tests}
To validate the statistical significance of our results, we performed the Friedman statistical test by using $F_M(\tilde{x})$ and $F_A(\tilde{x})$ of the multiple runs over five seeds of all our tested methods, i.e., NACHOS, NACHOS\_NP, NACHOS\_NC, and EDANAS, for each dataset and reported results in Table \ref{tab:friedman_results}. The results shows that there are statistic differences among the methods, hence the competitive performance of NACHOS is statistically validated.

\begin{table}[ht]
\centering
\caption{Friedman test results for each dataset and objective. The higher $\chi^2$, the more different the results of the methods based on their rankings and the smaller the p-value (typically $<0.05$), the more statistically significant the $\chi^2$ statistic.}
\label{tab:friedman_results}
\begin{tabular}{lcc}
\toprule
\textbf{Dataset \& Objective} & \textbf{Chi-squared ($\chi^2$)} & \textbf{p-value} \\
\midrule
SVHN ($F_M(\tilde{x})$)         & 9.9600  & 0.0189 \\
SVHN ($F_A(\tilde{x})$)         & 15.0000 & 0.0018 \\
CIFAR-10 ($F_M(\tilde{x})$)     & 14.0400 & 0.0029 \\
CIFAR-10 ($F_A(\tilde{x})$)     & 13.5600 & 0.0036 \\
CINIC-10 ($F_M(\tilde{x})$)     & 10.5625 & 0.0143 \\
CINIC-10 ($F_A(\tilde{x})$)     & 13.0408 & 0.0045 \\
Imagenette ($F_M(\tilde{x})$)   & 15.0000 & 0.0018 \\
Imagenette ($F_A(\tilde{x})$)   & 15.0000 & 0.0018 \\
\bottomrule
\end{tabular}
\end{table}

\section{Conclusions}
\label{sct:conclusions}

In this work, we have presented Neural Architecture Search for Hardware Constrained Early Exit Neural Networks (NACHOS), the first NAS framework for the design of optimal EENNs satisfying a constraint on the number of Multiply and Accumulate (MAC) operations performed by the EENNs at inference time. The results show that NACHOS is able to select a set of Pareto Optimal Solutions competitive with the other state-of-the-art EENNs. Additionally, we investigated the effectiveness of the novel regularizations introduced in the training of the EENNs proposed in NACHOS. Future works will examine different OFAs, early-exit schemes, optimization techniques, application scenarios, and the combination of NACHOS with other AutoML techniques.
In particular, we stress that one limitation of the current research is that it primarily focuses on mobile devices, optimizing neural networks with constraints in the range of millions of parameters and MACs. 
Future research directions could focus on further optimization of neural networks by incorporating hardware accelerator architectures and compiler mappings directly into the design space. This could lead to the development of models that are efficient enough to be deployed on MCUs and in general to scenarios with very severe constraints.

%-----------------------------------------------------------------------------
% BIBLIOGRAPHY
%-----------------------------------------------------------------------------
\bibliographystyle{IEEEtran}
\bibliography{IEEEabrv,main2.bib}

\end{document}